%% file: main.tex
\documentclass[10pt,twocolumn,letterpaper]{article}

\usepackage[pagenumbers]{cvpr} 

\input{preamble}

\definecolor{cvprblue}{rgb}{0.21,0.49,0.74}
\usepackage[pagebackref,breaklinks,colorlinks,allcolors=cvprblue]{hyperref}
\usepackage{multirow}
\usepackage[table]{xcolor}
\usepackage{threeparttable}
\usepackage[most]{tcolorbox}
\usepackage{cuted}

\def\paperID{20314} 
\def\confName{CVPR}
\def\confYear{2026}

\title{ReLaX: Reasoning with Latent Exploration for Large Reasoning Models}

\usepackage{marvosym}
\author{
Shimin Zhang$^{1*}$\quad
Xianwei Chen$^{1*}$\quad
Yufan Shen$^{2*}$\quad
Ziyuan Ye$^{1}$\quad
Jibin Wu$^{1^\dagger}$\\
$^{1}$Hong Kong Polytechnic University \quad
$^{2}$Shanghai Artificial Intelligence Laboratory\\[0.5em]
{\small * Equal contribution \quad $\dagger$ \ Corresponding author: \texttt{jibin.wu@polyu.edu.hk}}
}

\begin{document}
\maketitle
\input{sec/0_abstract}    
\input{sec/1_intro}

\input{sec/2_preliminary}
\input{sec/3_method}

\input{sec/4_experiments}
\input{sec/5_conclusion}
{
    \small
    \bibliographystyle{ieeenat_fullname}
    \bibliography{main}
}

\input{sec/X_suppl}

\end{document}

%% file: preamble.tex
\usepackage{tabularx}
\usepackage[most,skins,theorems]{tcolorbox}
\tcbset{
  aibox/.style={
    width=\linewidth,
    top=7pt,
    bottom=2pt,
    colback=blue!6!white,
    colframe=black,
    colbacktitle=black,
    enhanced,
    center,
    attach boxed title to top left={yshift=-0.1in,xshift=0.15in},
    boxed title style={boxrule=0pt,colframe=white,},
  }
}
\newtcolorbox{AIbox}[2][]{aibox,title=#2,#1}

\usepackage[linesnumbered,ruled,vlined]{algorithm2e}



%% file: sec/0_abstract.tex
\begin{abstract}
Reinforcement Learning with Verifiable Rewards (RLVR) has recently demonstrated remarkable potential in enhancing the reasoning capability of Large Reasoning Models (LRMs). However, RLVR often drives the policy toward over-determinism, resulting in ineffective exploration and premature policy convergence.
While promoting token-level diversity has shown promise in mitigating entropy collapse, we argue that the latent dynamics underlying token generation encode a far richer computational structure for steering policy optimization toward a more effective exploration–exploitation tradeoff.
To enable tractable analysis and intervention of the latent dynamics of LRMs,  we leverage Koopman operator theory to obtain a linearized representation of their hidden state dynamics. This enables us to introduce \textbf{D}ynamic \textbf{S}pectral \textbf{D}ispersion (\textbf{DSD}),
a new metric to quantify the heterogeneity of the model's latent dynamics, serving as a direct indicator of policy exploration.
Building upon these foundations, we propose \textbf{Re}asoning with \textbf{La}tent e\textbf{X}ploration (\textbf{ReLaX}), a framework that explicitly incorporates latent dynamics to regulate exploration and exploitation during policy optimization. 
Comprehensive experiments across a wide range of multimodal and text-only reasoning benchmarks show that ReLaX consistently incentivizes reasoning capability and outperforms existing token-level methods. Our project is available at \href{https://github.com/ZhangShimin1/ReLaX}{https://github.com/ZhangShimin1/ReLaX}.
\end{abstract}

%% file: sec/1_intro.tex
\section{Introduction}
\label{sec:intro}

\begin{figure}[htbp]
  \centering
  \begin{subfigure}[t]{0.23\textwidth}
    \centering
    \includegraphics[width=\linewidth]{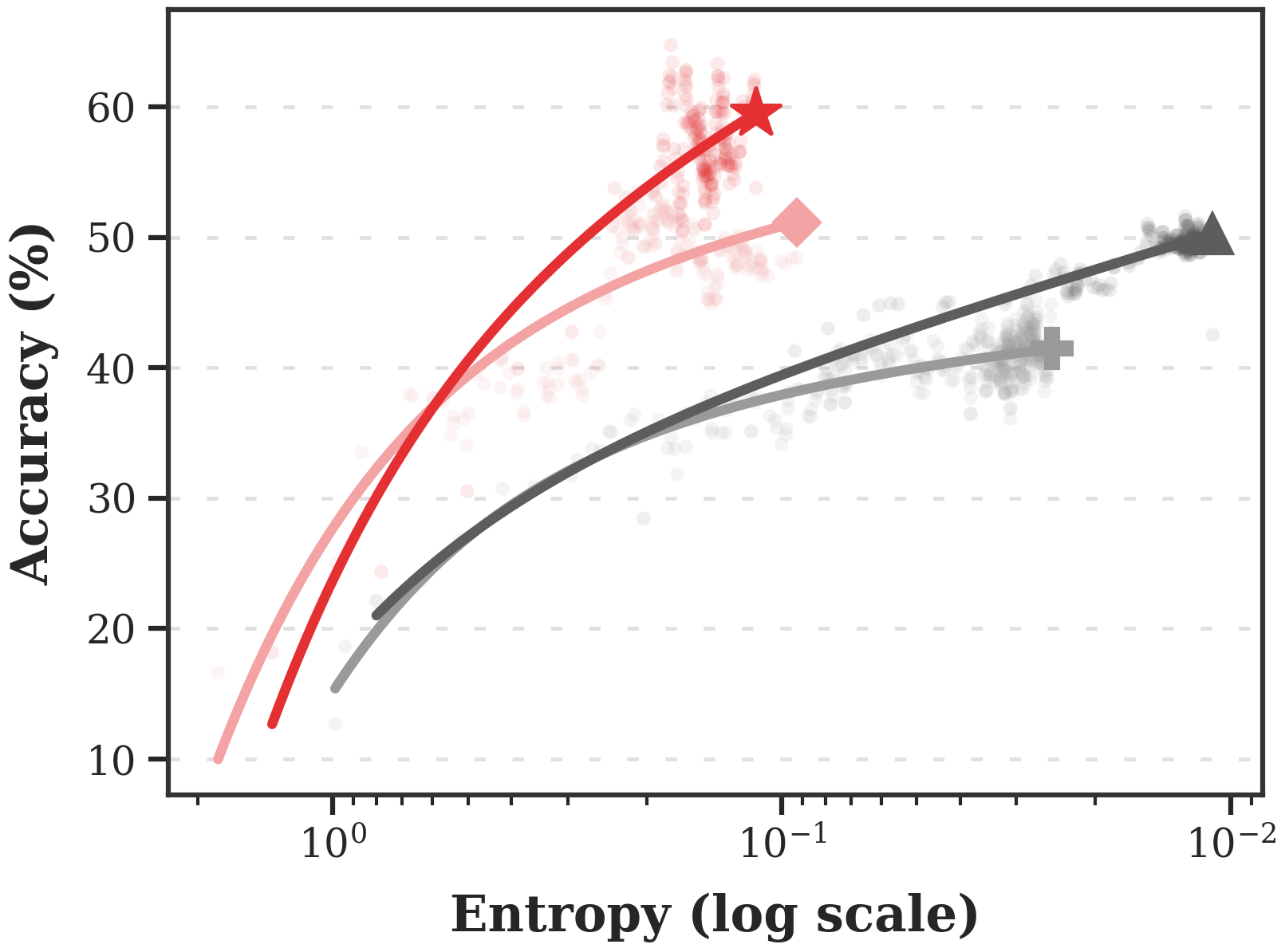}
    \caption{LLMs (Text-only)}
    \label{fig1:a}
  \end{subfigure}
  \hfill
  \begin{subfigure}[t]{0.23\textwidth}
    \centering
    \includegraphics[width=\linewidth]{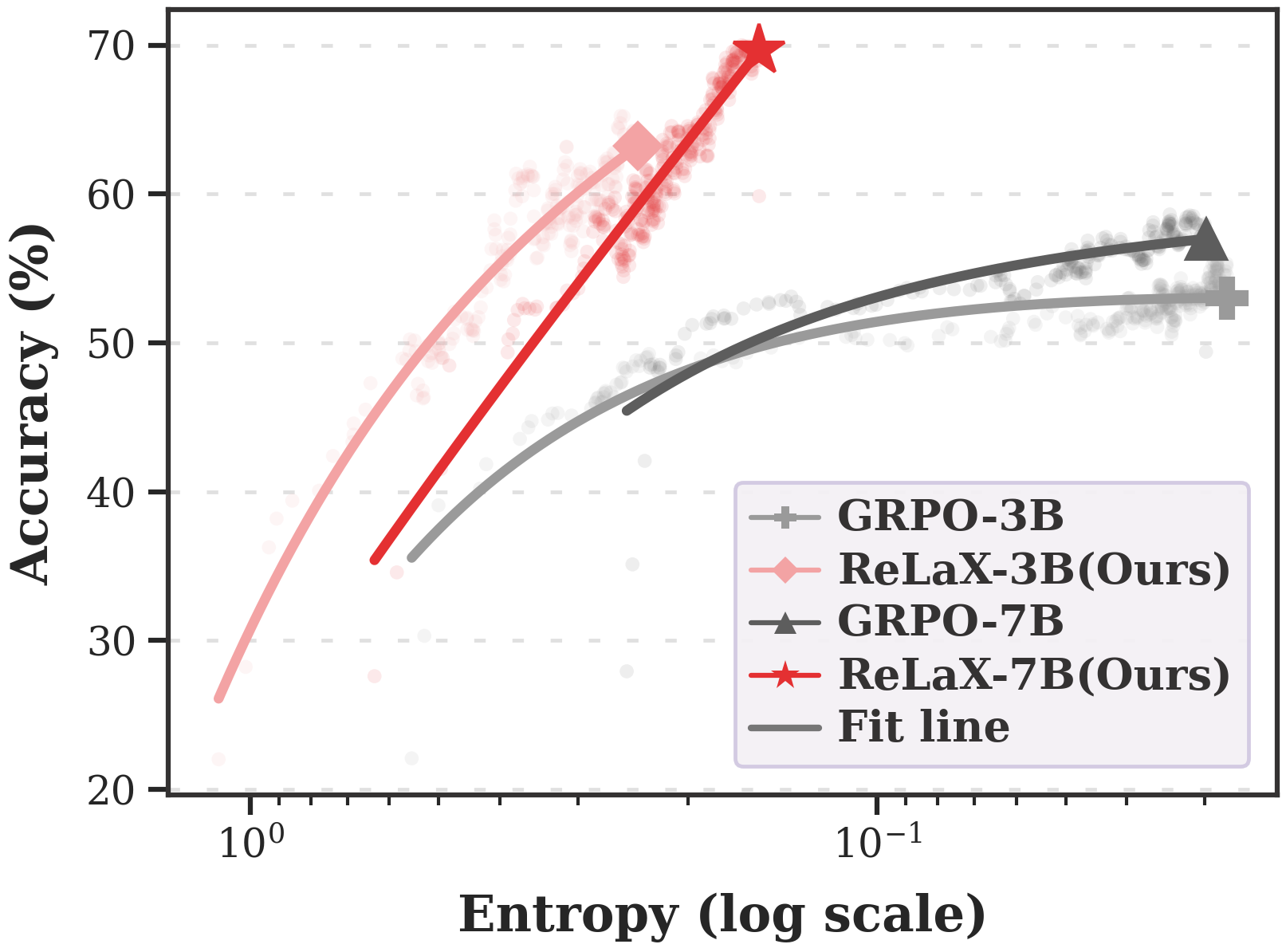}
    \caption{MLLMs (Vision-language)}
    \label{fig1:b}
  \end{subfigure}

  \caption{Empirical relationship between policy performance $\mathcal{R}$ and token-level entropy $H$ during RLVR training with (a) text-only LLMs and (b) vision-language models (VLMs). Each scatter denotes a single training step, the solid curve fitted by $\mathcal{R} = -a \cdot \exp(H) + b$~\cite{cui2025entropy}. 
  }
  \label{fig:performance-bottleneck}
\end{figure}

\begin{figure*}[ht]
    \centering
    \includegraphics[width=.95\linewidth]{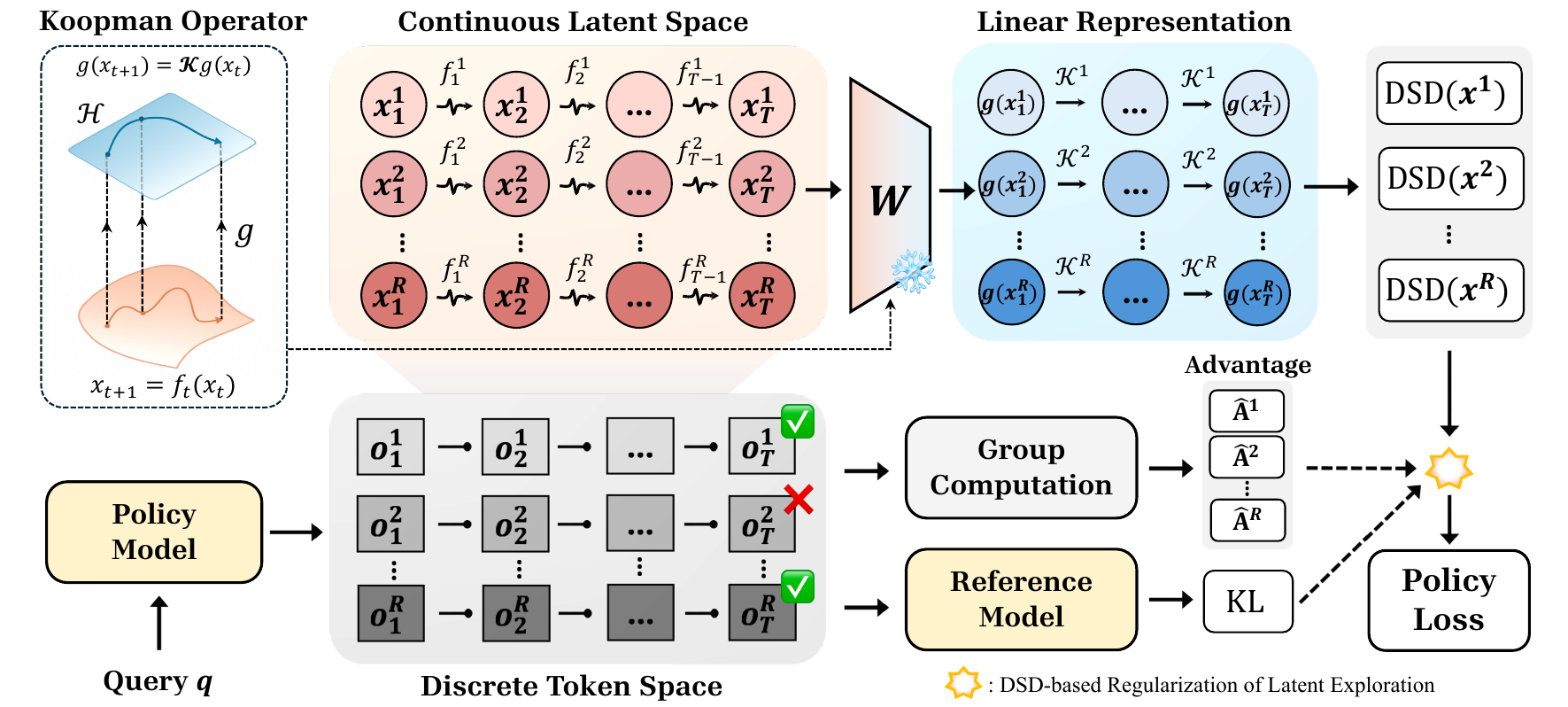}
    \caption{\textbf{Overview of ReLaX.} Grounded in Koopman operator theory (upper left), ReLaX employs a neural Koopman dictionary (frozen after one step of learning) during policy optimization to linearize the latent dynamics of last-layer hidden states. This transformation allows us to assess the flexibility of policy's internal computations through the proposed DSD. The DSD score for each trajectory is subsequently integrated into the GRPO objective, mitigating computational rigidity and enabling a more effective exploration–exploitation tradeoff.}
    \label{fig:framework}
\end{figure*}

Scalable and verifiable reasoning stands as the foundational capability separating current foundation models from artificial general intelligence.
Recent progress~\cite{jaech2024openai,guo2025deepseek,comanici2025gemini,peng2025skywork} in RLVR has emerged as an effective paradigm for enhancing the capability of Large Language Models (LLMs) and Multimodal LLMs (MLLMs) on complex reasoning tasks.
Whereas, without explicit intervention, reinforcement learning (RL) progressively concentrates the policy distribution and reduces entropy, confining the policy gradient within a narrow subspace~\cite{agarwal2021theory, shen2025entropy}. 
The sparse reward in RLVR further exacerbates this \textit{entropy collapse} problem, causing the policy to over-exploit prematurely, thereby inhibiting adequate exploration and ultimately leading to suboptimal performance.
This bottleneck is supported by empirical evidence from Group Relative Policy Optimization (GRPO)~\cite{shao2024deepseekmath,zeng2025simplerl}, which reveals an exponential relationship between policy entropy $H$ and reward $\mathcal{R}$: $\mathcal{R} = -a \cdot \exp(H) + b$~\cite{cui2025entropy} (see grey lines in Figure~\ref{fig:performance-bottleneck}). To date, the community has widely recognized the exploration–exploitation tradeoff~\cite{sutton1988learning} as the fundamental challenge in scaling RL to achieve improved reasoning performance~\cite{yue2025does,wu2025invisible,wang2025stabilizing,li2025know, hu2025diversity,deng2025decomposing}. 

Existing endeavors (with a review in the Supplementary Material) mainly focus on raising token-level entropy, including reshaping the reward~\cite{cheng2025reasoning,hu2025diversity,chen2025pass}, redesigning policy objectives with entropy-based regularization~\cite{yao2025diversity, lei2025revisiting}, and heuristically anchoring salient tokens to locally elevate the stochasticity~\cite{zheng2025first,cui2025entropy,li2025cure,wang2025beyond,yu2025dapo,yang2025dcpo, wang2025emergent}.  Despite recent progress, the objective of maintaining higher token-level entropy inherently conflicts with the tendency of RL, which naturally gravitates toward deterministic, low-entropy policies. Moreover, mainstream MLLMs~\cite{wang2024qwen2,bai2025qwen2,xu2025qwen3} exhibit a pronounced misalignment between cross-modal internal computations and unimodal, text-centric outputs, making it difficult for token-level feedback to accurately reflect the underlying multimodal processing~\cite{sun2024generative,xie2025showo}. Taken together, these factors render existing methods inefficient and limit their generalizability when applied to MLLMs.

In this paper, we argue that entropy collapse is the superficial symptom of a deeper pathology: under RLVR, the internal computations that govern token generation gradually lose flexibility and converge into overly rigid patterns. These computations are instantiated through the high-dimensional latent dynamics of hidden states that embed far richer and more stable inductive biases than what is observable in the discrete and sensitive token space.
The key to harnessing the continuous latent dynamic of LRMs lies in finding an appropriate representation to make it analytically tractable. Modern Koopman operator theory~\cite{brunton2021modern} provides a powerful framework to represent the nonlinear dynamics of a model as linear evolution in an infinite-dimensional space of observables, with the Koopman dictionary governing the functional coordinates of this representation.
Building upon this foundation, we introduce a novel metric, \textbf{Dynamic Spectral Dispersion (DSD)}, to characterize the flexibility of internal computations in LRMs by quantifying the degree of heterogeneity exhibited in their underlying latent dynamics.
We then propose a latent dynamics aware framework for RLVR, \textbf{ReLaX} (\textbf{Re}asoning with \textbf{La}tent e\textbf{X}ploration), that employs DSD-based regularization to counteract the rigid internal computation and facilitate effective exploration-exploitation tradeoff.
Our main contributions are summarized as follows:
\begin{itemize}
\item We propose DSD, a metric that captures the heterogeneity of latent reasoning dynamics in LRMs, providing a more fundamental characterization of policy exploration by probing the model’s internal computational processes rather than its surface-level token statistics.
\item Incorporating DSD into the objective of policy optimization, we introduce a novel framework, ReLaX, to effectively facilitate the exploration–exploitation tradeoff and mitigate the performance saturation of RLVR (see Figure~\ref{fig:performance-bottleneck}).
\item Extensive experiments demonstrate that ReLaX substantially advances LRM capabilities, outperforming token-level methods and setting new state-of-the-art results on $7$ multimodal and $6$ text-only reasoning benchmarks. These findings indicate that ReLaX opens a promising direction for advancing reasoning capabilities by transcending token-space interventions and guiding exploration within the more expressive and computationally meaningful latent space.
\end{itemize}

%% file: sec/2_preliminary.tex
\section{Preliminary}
\label{sec:preliminary}

\subsection{Problem Formulation}
The fine-tuning of foundation models for reasoning tasks can be formulated as a reinforcement learning problem driven by \emph{verifiable rewards}.
Let $\pi_\theta(o|q)$ denote the policy model parameterized by $\theta$, which generates a reasoning trajectory 
$o = (o_0, o_1, \dots, o_T)$ conditioned on a prompt $q \sim \mathcal{D}$, 
where $\mathcal{D}$ denotes the distribution of input prompts. 
The trajectory length is constrained by a maximum context length limit $T \le L_{\max}$.

Each generated trajectory $o$ is evaluated by a \emph{verifier} that produces a scalar-valued $\mathrm{reward}(q, o)$.
The verifier provides an objective and automatically checkable signal 
(e.g., correctness of a mathematical derivation or successful code execution), 
thus avoiding subjective biases and reward hacking issues in preference-based RL. 
The training objective of RLVR is to maximize the expected reward under the policy distribution:
\begin{equation}
\max_{\theta} \; \mathcal J(\theta) = 
\mathbb{E}_{q \sim \mathcal{D},\, o \sim \pi_\theta(\cdot|q)}
\big[ \mathrm{reward}(q, o) \big],
\label{eq:rlvr}
\end{equation}
subject to $|o| \le L_{\max}$.

\subsection{Group Relative Policy Optimization}
To efficiently optimize the RLVR objective, GRPO serves as a group-based variant of PPO, stabilizing policy improvement by normalizing rewards within each prompt group.
For each prompt $q$, GRPO samples a group of $R$ responses $\{o^i\}_{i=1}^{r}$ and estimates a group-relative advantage for each trajectory:
\begin{equation}
\hat A^i = \frac{\mathrm{reward}(q,o^i) - \mathrm{mean}\big[\mathrm{reward}(q,\{o^i\}_{i=1}^{R})\big]}
{\mathrm{std}\big[\mathrm{reward}(q,\{o^i\}_{i=1}^{R})\big]}.
\label{eq:grpo-adv}
\end{equation}
To handle off-policy data and constrain the optimization step, GRPO adopts the PPO-style clipped surrogate objective:
\begin{equation}
\begin{array}{rl}
\displaystyle\mathcal{J}_{\mathrm{GRPO}}(\theta) & \displaystyle=\mathbb{E}_{q \sim \mathcal{D}, \{o^i\}_{i=1}^{R} \sim \pi_{\theta_{\mathrm{old}}}(\cdot \mid q)}  \\[1em]
& \displaystyle\Biggr[
\frac{1}{R}\sum_{i=1}^{R}
\min\left(
\frac{\pi_\theta(o^i \mid q)}{\pi_{\theta_{\mathrm{old}}}(o^i \mid q)}\,A^i, \right.\\[1.5em]
& \displaystyle\left. \mathrm{clip}\!\left(
\frac{\pi_\theta(o^i \mid q)}{\pi_{\theta_{\mathrm{old}}}(o^i \mid q)},\, 1-\epsilon,\, 1+\epsilon
\right) A^i
\right) \Biggr].
\end{array}
\label{eq:grpo-loss}
\end{equation}
To constrain policy updates and ensure stability, a KL-divergence regularization $\mathcal{L}_{\mathrm{KL}}= D_{\mathrm{KL}}\!\left(\pi_\theta\|\,\pi_{\mathrm{ref}}\right)$ scaled by $\beta$ is optionally applied to penalize large deviations from the initial policy.
By leveraging group-relative normalization, GRPO provides a stable and scale-invariant learning signal, effectively reducing variance and improving sample efficiency in reasoning-oriented policy updates.

\subsection{Koopman Operator Theory and Dynamic Mode Decomposition}
The Koopman operator provides a rigorous framework for representing a nonlinear dynamical system 
within an infinite-dimensional Hilbert space $\mathcal{H}$. For a discrete-time system governed by $x_{t+1}=f_{t}(x_{t})$, the Koopman operator $\mathcal{K}$ acts on observables $g\in \mathcal{H}$ (Koopman dictionary) as:
\begin{equation}
    [\mathcal{K}g](x_t):=g(f_t(x_t)) = g(x_{t+1}).
\end{equation}
Building upon Koopman operator theory, the Dynamic Mode Decomposition (DMD)~\cite{schmid2022dynamic} serves as a data-driven finite-dimensional approximation of $\mathcal{K}$. 
Consider a trajectory with consecutive states $\mathcal{V}=\{g(x_0), g(x_1) \dots, g(x_{t-1})\}$ and their successors $\mathcal{V}^+=\{g(x_1), g(x_2) \dots, g(x_{t})\}$,
$\mathcal{K}$ can be
estimated by solving a least-squares problem:
\begin{equation}
\label{eq:koop-approx}
\mathcal{K} = \arg\min_{\mathcal{K}} \left\|\mathcal{V}^+ - \mathcal{K}\mathcal{V} \right\|_F^2 = \mathcal{V}^+ \mathcal{V}^{\dagger},
\end{equation}
where $\mathcal{V}^{\dagger}$ denotes the pseudoinverse of $\mathcal{V}$.

Spectral analysis on $\mathcal{K}$ through its eigenvalues enables the characterization of underlying nonlinear dynamics.
However, the discretization in DMD often introduces spurious eigenvalues when applied to complex systems with rich continuous spectra, leading to the loss of critical dynamical modes~\cite{li2017extended,williams2015data}.
Recent theoretical advances, particularly ResDMD ~\cite{colbrook2024rigorous}, address this challenge by filtering out poorly convergent spectral components, thereby enabling a more accurate and reliable characterization of complex dynamics. This approach has also shown promise for analyzing the hidden state dynamics of LLMs ~\cite{zhang2025koopstd}. 

%% file: sec/3_method.tex
\section{Proposed Methodology}
\label{sec:method}
This section gives a comprehensive presentation of the proposed ReLaX.
First, building on Koopman operator theory, Sec.~\ref{sec:dsd} introduces DSD as a principled metric for capturing the heterogeneity of latent dynamics (Fig.~\ref{fig:framework}, upper). Second, Sec.~\ref{sec:koop_dict} describes how we we construct a reliable and accurate linear representation to support calculating DSD for LRMs. Finally, Sec.~\ref{sec:relax} details how DSD is incorporated into GRPO to achieve an effective exploration–exploitation tradeoff (Fig.~\ref{fig:framework}, lower right).

\subsection{Dynamic Spectral Dispersion (DSD)}
\label{sec:dsd}

The explicit Chain-of-Thought (CoT) in LRMs is decoded from latent hidden representations evolving under:
\begin{equation}
    x_t = \mathcal{F}(x_{t-1}, \omega_t), \quad \omega_t \sim \mathcal{P}_\omega,
\end{equation}
where $\mathcal{F}$ defines a stochastic nonlinear dynamical system over the latent state space. The stochasticity $\omega_t$ (e.g., temperature, top-$p$, and top-$k$ sampling) not only determines the decoded CoT, but also injects perturbations into the latent trajectory. The extent to which these perturbations induce effective exploration depends on the richness of the intrinsic hidden state dynamics, which determines whether stochastic inputs can be translated into diverse latent trajectories.

However, the nonlinear and high-dimensional nature of latent dynamics presents substantial challenges for effectively and accurately capturing their underlying dynamics. The Koopman operator offers a rigorous framework for analyzing nonlinear dynamics by embedding the latent dynamics into an infinite-dimensional function space, where their evolution is governed by a linear operator $\mathcal{K}$. The spectral decomposition of $\mathcal{K}$ reveals fundamental dynamical modes (e.g., growth, decay, and oscillation) that govern the structure of the resulting reasoning trajectories.
Therefore, the distribution of spectral modes characterizes the intrinsic structure of the underlying dynamics: a concentrated spectrum implies degenerate, repetitive behavior, whereas a dispersed spectrum signifies a richer and more expressive dynamical system.

Leveraging this insight, we put forward DSD as an operational proxy for the policy model’s computational flexibility. Formally, given a sequence of hidden states $x \in \mathbb{R}^{T \times d}$ and its approximated Koopman operator $\mathcal{K}$, the DSD is defined as the variance of the Koopman eigenvalue magnitudes:
\begin{equation}
\label{eq:dsd}
    \mathrm{DSD}(x) = \operatorname{Var}(|\Lambda|),  \hspace{4pt}\mathrm{where} \hspace{4pt}\mathcal{K}\Phi=\Phi\Lambda .
\end{equation}
Higher DSD scores correspond to richer intrinsic dynamical spectra, indicating that the model avoids collapsing into rigid computational patterns. This dynamical flexibility enables stochastic perturbations to be effectively leveraged, giving rise to diverse latent trajectories that support sustained and efficient exploration during policy optimization. The evolution of latent trajectory in a low-dimensional space is visualized in the Supplementary Material.

\subsection{Koopman Dictionary Learning}
\label{sec:koop_dict}
A core challenge in applying the Koopman operator is selecting an appropriate function space in which the operator $\mathcal{K}$ can faithfully linearize the policy model’s latent dynamics—an especially difficult problem given the high dimensionality and strong nonlinearity of LRMs. To address this, we adopt ResKoopNet~\cite{xu2025reskoopnet}, an extension of ResDMD that learns a neural Koopman dictionary. This enables a more accurate and numerically stable approximation of the Koopman operator and its spectrum, which is essential for reliable DSD computation.

Specifically, the Koopman observables $g$ are parameterized by a single linear layer $W$ followed by a sigmoid activation $\sigma(\cdot)$:
\begin{equation}
    g(x) = \sigma(W x), 
    \quad W \in \mathbb{R}^{d \times m},
\end{equation}
where $m$ represents the dimensionality of the approximated Koopman operator.
The dictionary $W$ is optimized using latent trajectories $\{x_i\}_{i=1}^{B \times R}$ collected from the last hidden-layer of initial policy, where $B$ denotes the policy training batch size.
The optimization objective for $W$ is to minimize the spectral residual of the Koopman operator:
\begin{equation}
\label{eq:reskoopnet}
    W = \arg\min \frac{1}{BR}\,
    \,
    \left\| \big(\mathcal{V}^+ - \mathcal{K}\mathcal{V}\big) \Phi \,
    \right\|_{F}^{2},
\end{equation}
where $\Phi$ denotes the eigenvectors of $\mathcal{K}$. 
Once optimized, $W$ is frozen during subsequent policy training to ensure a consistent function space for portraying policy latent dynamics throughout optimization.
A more detailed presentation related to Koopman dictionary learning is provided in the Supplementary Material.

\subsection{ReLaX}
\label{sec:relax}
DSD characterizes the policy exploration from a computational perspective and supports gradient propagation, providing a principled basis for steering policy optimization toward an effective exploration–exploitation tradeoff.
Formally, for hidden states $\{x^i\}_{i=1}^{R}$ correspond to group responses $\{o^i\}_{i=1}^{R}$, we define a sequence-level regularization term $\mathcal{L}_{\mathrm{xp}}$ associated with the corresponding DSD scores as:
\begin{equation}
    \mathcal{L}_{\mathrm{xp}} 
    = \log\!\left(
        \frac{1}{R}
        \sum_{i=1}^{R} 
        \exp\!\big(-\mathrm{DSD}(x^i)\big)
    \right),
\end{equation}
where the log-mean-exp computation smooths DSD numerically, enhancing gradient stability during optimization. 

While this regularization alleviates computational rigidity by discouraging the latent dynamics from converging to a homogeneous mode, excessive exploration can undermine the necessary exploitation. Accordingly, ReLaX integrates two control mechanisms to maintain latent exploration at an appropriate operating level.
First, to ensure that exploration occurs within a meaningful subspace of the latent dynamics, we weight the DSD scores by the advantages truncated to their positive part.  This design constrains the policy model to become more flexible only along trajectories that yield positive reward, preventing uninformative or detrimental exploration. The resulting advantage-shaped regularization is expressed as:
\begin{equation} \tilde{\mathcal{L}}_{\text{xp}} = \log\!\left( \frac{1}{R} \sum_{i=1}^{R} \exp\!\big(- \mathrm{clip}(\hat A^i,0) \cdot \mathrm{DSD}(x^i)\big) \right).
\label{equ:adv-shaped}
\end{equation}
Nevertheless, an over-dispersion of dynamic spectral modes within positive trajectories still induces instability. KL regularization acts as an elastic constraint to stabilize training, albeit at the cost of slower convergence~\cite{yu2025dapo, cui2025entropy, zheng2025first, yang2025dcpo}. We compormisely apply an adaptive KL regularization to softly constrain policy updates for trajectories that exhibit excessive dynamic divergence, while allowing those with remaining exploration potential to proceed freely. The overall objective of ReLaX is thus formulated as:
\begin{equation}
\mathcal{J}(\theta) =\mathcal{J}_{\mathrm{GRPO}}(\theta) + \alpha \hspace{1pt} \tilde{\mathcal{L}}_{\text{xp}} +
\beta\hspace{1pt}\sum_{i}^{\mathcal{I}} D_{\mathrm{KL}}\!\left(\pi_\theta(o^i)\,\|\,\pi_{\mathrm{ref}}(o^i)\right),
\label{eq:relax_obj}
\end{equation}
where $\alpha$ controls the strength of regularization for latent exploration, and $\mathcal{I}=\{\, i \mid \mathrm{DSD}(x^i) > \xi \,\}$ denotes the subset of trajectories whose DSD scores exceed a threshold $\xi$, thereby requiring KL penalization. A complete pseudocode of ReLaX is included in the Supplementary Material.

%% file: sec/4_experiments.tex
\begin{table*}[ht]
\centering
\small
\setlength{\tabcolsep}{5pt}
\renewcommand{\arraystretch}{1.15}
\begin{tabular}{lcccccccc}
\toprule
\multirow{2}{*}{\textbf{Model}} & 
\multicolumn{1}{c}{\textbf{MathVista}} & 
\multicolumn{1}{c}{\textbf{MathVerse}} & 
\multicolumn{1}{c}{\textbf{MathVision}} & 
\multirow{1}{*}{\textbf{DynaMath}} & 
\multirow{1}{*}{\textbf{MMMU}} & 
\multirow{1}{*}{\textbf{MMStar}} & 
\multirow{1}{*}{\textbf{EMMA}} & 
\multirow{2}{*}{\textbf{Average}} \\
 &  \textit{testmini} & \textit{testmini} & \textit{test} & \textit{overall} & \textit{val} & \textit{overall} & \textit{overall} \\
\midrule
\multicolumn{9}{c}{\textbf{General Multimodal LLMs}} \\
\midrule
Qwen2-VL-7B~\cite{wang2024qwen2}  & 58.2 & 19.7 & 16.3 & 42.1 & 54.1 & 60.7 & 20.2 & 38.8 \\
Qwen2.5-VL-3B~\cite{bai2025qwen2}  & 62.3 & 33.5$^{\dagger}$ & 21.2 & 40.0$^{\dagger}$ & 46.3 & 55.9 & 19.2$^{\dagger}$ & 39.8 \\
Qwen2.5-VL-7B~\cite{bai2025qwen2}  & 68.2 & 49.2 & 25.1 & 53.2 & 54.3 & 63.9 & 21.5 & 47.9 \\
Qwen2.5-VL-72B~\cite{bai2025qwen2}  & 74.8 & 57.6 & 38.1 & - & 70.2 & 70.8 & - & - \\
Intern2-VL-8B~\cite{chen2024expanding}  & 58.3 & 22.8 & 17.4 & 39.7 & 51.2 & 61.5 & 19.8 & 38.7 \\
Intern2.5-VL-8B~\cite{chen2024expanding} & 64.4 & 39.5 & 19.7 & - & 56.0 & - & 20.6 & - \\
Llava-OV-7B~\cite{li2025llavaonevision} & 63.2 & 26.2 & 18.5 & - & 48.8 & - & 18.3 & - \\
Kimi-VL-16B~\cite{team2025kimi} & 68.7 & 44.9 & 21.4 & - & 55.7 & - & - & - \\
\midrule
\multicolumn{9}{c}{\textbf{Reasoning Multimodal LLMs}} \\
\midrule
MM-Eureka-7B~\cite{meng2025mm}  & 73.0 & 50.3 & 26.9 & 54.4$^{\dagger}$ & 55.2 & \textcolor{RoyalBlue}{64.3$^{\dagger}$} & 23.5 & 49.7 \\
MM-Eureka-8B~\cite{meng2025mm}  & 67.1 & 40.4 & 22.2 & - & 49.2 & - & 21.5 & - \\
Vision-R1-7B~\cite{huang2025vision}  & 73.5 & 52.4 & 27.2 & 52.0$^{\dagger}$ & 54.7 & 60.9$^{\dagger}$ & 22.4 & 49.0 \\
R1-VL-7B~\cite{zhang2025r1} & 63.5 & 40.0 & 24.7 & 45.2 & 44.5 & 60.0 & 8.3 & 40.9 \\
OpenVLThinker-7B~\cite{deng2025openvlthinker} & 70.2 & 47.9 & 25.3 & 38.6$^{\dagger}$ & 52.5 & 56.3$^{\dagger}$ & 26.6 & 45.3 \\
VL-Rethinker-7B~\cite{wang2025vlrethinker} & 74.9 & 54.2 & \textcolor{RoyalBlue}{32.3} & \textcolor{RoyalBlue}{55.2$^{\dagger}$} & 56.7 & 64.2$^{\dagger}$ & \textcolor{RoyalBlue}{29.7} & \textcolor{RoyalBlue}{52.5} \\
SRPO-7B~\cite{jiang2023understanding} & \textcolor{RoyalBlue}{75.8} & \textcolor{BrickRed}{55.8} & \textcolor{BrickRed}{32.9} & - & \textcolor{RoyalBlue}{57.1} & - & 29.6 & - \\
\midrule
\rowcolor{blue!8}
\textbf{\textcolor{black}{ReLaX-VL-3B (Ours)}} & 70.7 & 46.2 & 27.6 & 52.2 & 52.2 & 60.7 & 26.9 & \textbf{48.1} \\
\rowcolor{blue!8}
\textbf{\textcolor{black}{ReLaX-VL-7B (Ours)}} & \textcolor{BrickRed}{77.1} & \textcolor{RoyalBlue}{55.7} & 30.2 & \textcolor{BrickRed}{55.9} & \textcolor{BrickRed}{57.4} & \textcolor{BrickRed}{65.5} & \textcolor{BrickRed}{30.6} & \textcolor{BrickRed}{\textbf{53.2}}  \\
\bottomrule
\end{tabular}
\caption{Comparison of VLM performance (mean@1 accuracy) across multiple multimodal reasoning benchmarks. For 7B scale LRMs, the top-performing and runner-up results of VLMs within each column are marked in \textcolor{BrickRed}{red} and \textcolor{RoyalBlue}{blue}, respectively. $^{\dagger}$ indicates our reproduced results using publicly available models and standard evaluation code. “–” denotes missing results due to unavailable models.}
\label{tab:multimodal_comparison}
\end{table*}
\section{Experiments}
\label{sec:exp}

\begin{figure*}
    \centering
    \includegraphics[width=1.0\linewidth]{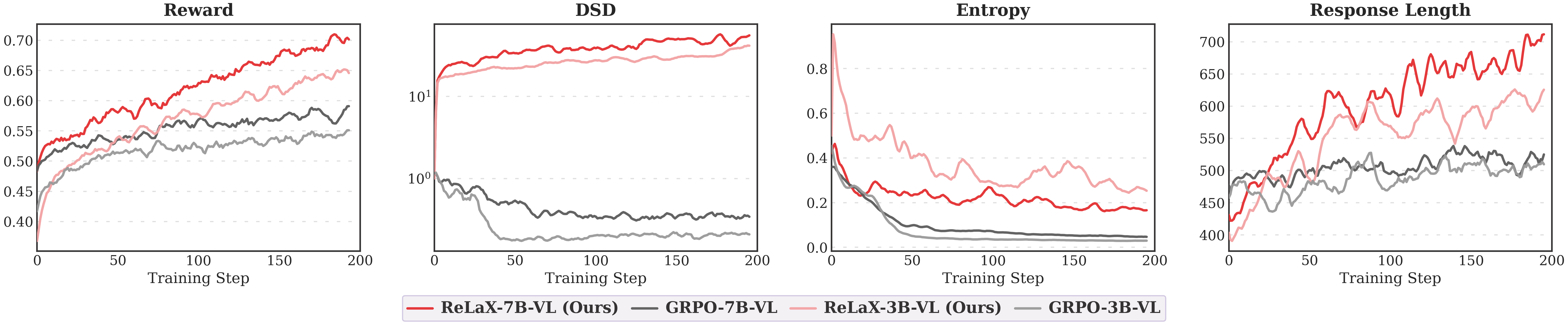}
    \caption{Comparison of training dynamics for \textit{Reward}, \textit{DSD}, \textit{Entropy}, and \textit{Response Length} under ReLaX (red) and vanilla GRPO (gray) on Qwen2.5-VL-Instruct at the 3B and 7B scales.}
    \label{fig:ReLaX-training-dynamics}
\end{figure*}

\begin{table*}[ht]
\centering
\small
\setlength{\tabcolsep}{5pt}
\renewcommand{\arraystretch}{1.15}
\begin{tabular}{lcccccccccc}
\toprule
\multirow{2}{*}{\textbf{Model}} & 
\multirow{2}{*}{\textbf{Size}} & 
\multicolumn{1}{c}{\textbf{MATH500}} & 
\multicolumn{1}{c}{\textbf{Minerva}} & 
\multicolumn{1}{c}{\textbf{AMC22}} & 
\multicolumn{1}{c}{\textbf{AMC23}} & 
\multicolumn{1}{c}{\textbf{AIME24}} & 
\multicolumn{1}{c}{\textbf{AIME25}} & 
\multirow{2}{*}{\textbf{Average}} \\
\cmidrule(lr){3-8}
 &  & \textit{Mean@1} & \textit{Mean@1} & \textit{Mean@32} & \textit{Mean@32} & \textit{Mean@32} & \textit{Mean@32} &  \\
\midrule
Qwen2.5-base & 3B & 18.4 & 2.6 & 12.5$^{\dagger}$ & 7.0 & 0.6 & 0.3 & 5.9 \\
\quad +Vanilla GRPO\cite{shao2024deepseekmath} & 3B & 69.2 & \textcolor{RoyalBlue}{23.5$^{\dagger}$} & \textcolor{RoyalBlue}{25.9$^{\dagger}$} & 51.6 & 7.5 & 4.5 & \textcolor{RoyalBlue}{26.0} \\
\quad +DCPO\cite{yang2025dcpo} & 3B & \textcolor{RoyalBlue}{71.2} & - & - & \textcolor{RoyalBlue}{55.8} & \textcolor{RoyalBlue}{7.5} & \textcolor{RoyalBlue}{4.7} & - \\
\rowcolor{blue!8}
\quad \textbf{\textcolor{black}{+ReLaX (Ours)}} & 3B & \textcolor{BrickRed}{71.6} & \textcolor{BrickRed}{26.1} & \textcolor{BrickRed}{44.8} & \textcolor{BrickRed}{63.5} & \textcolor{BrickRed}{8.8} & \textcolor{BrickRed}{6.2} & \textcolor{BrickRed}{\textbf{31.6}} \\
\midrule
Qwen2.5-base & 7B & 64.6 & 5.7 & 21.4$^{\dagger}$ & 30.0 & 0.3 & 0.1 & 17.4 \\
\quad +SimpleRL\cite{zeng2025simplerl} & 7B & 78.2 & 38.6 & 39.5$^{\dagger}$ & 62.5 & 15.6 & 8.9$^{\dagger}$ & 34.8 \\
\quad +DAPO\cite{yu2025dapo} & 7B & 77.8 & 35.3 & - & 60.0 & 18.1 & 11.5 & - \\
\quad +KL-Cov\cite{cui2025entropy} & 7B & \textcolor{RoyalBlue}{80.8} & 38.2 & - & 61.4 & \textcolor{RoyalBlue}{22.6} & 12.9 & - \\
\quad +HICRA\cite{wang2025emergent} & 7B & 80.2 & 38.6 & - & 55.1 & 18.8 & \textcolor{BrickRed}{14.8} & - \\
\quad +FR3E\cite{zheng2025first} & 7B & 79.0 & \textcolor{RoyalBlue}{39.0} & \textcolor{RoyalBlue}{49.2$^{\dagger}$} & \textcolor{RoyalBlue}{67.5} & \textcolor{BrickRed}{25.2} & \textcolor{BrickRed}{14.8}$^{\dagger}$ & \textcolor{RoyalBlue}{39.2} \\
\rowcolor{blue!8}
\quad \textbf{\textcolor{black}{+ReLaX (Ours)}} & 7B & \textcolor{BrickRed}{82.4}  & \textcolor{BrickRed}{39.1}  & \textcolor{BrickRed}{65.4}  & \textcolor{BrickRed}{84.1}  & 19.7  & \textcolor{RoyalBlue}{13.8}  & \textcolor{BrickRed}{\textbf{43.5}} \\
\midrule
Qwen2.5-Math-base & 7B & 52.4 & 10.7 & 25.4$^{\dagger}$ & 52.2 & 16.6 & 6.3 & 23.4 \\
\quad +SimpleRL\cite{zeng2025simplerl} & 7B & 77.4 & 32.0 & 57.0$^{\dagger}$ & 60.8 & 26.7 & 9.3 & 37.6 \\
\quad +DAPO\cite{yu2025dapo} & 7B & 81.6 & 38.2 & - & 62.5 & 31.6 & 14.9 & - \\
\quad +DCPO\cite{yang2025dcpo} & 7B & \textcolor{RoyalBlue}{82.5} & - & - & \textcolor{RoyalBlue}{79.8} & \textcolor{BrickRed}{38.8} & 17.2 & - \\
\quad +R1-zero-Div\cite{yao2025diversity} & 7B & 77.2 & 31.6$^{\dagger}$ & 50.2$^{\dagger}$ & 58.0$^{\dagger}$ & 17.5$^{\dagger}$ & 10.0$^{\dagger}$ & 34.9 \\
\quad +FR3E\cite{zheng2025first} & 7B & 82.2 & \textcolor{RoyalBlue}{40.8} & \textcolor{RoyalBlue}{64.1$^{\dagger}$} & 67.5 & 26.7 & \textcolor{BrickRed}{18.1$^{\dagger}$} & \textcolor{RoyalBlue}{42.8} \\
\rowcolor{blue!8}
\quad \textbf{\textcolor{black}{+ReLaX (Ours)}} & 7B & \textcolor{BrickRed}{85.6}  & \textcolor{BrickRed}{43.4}  & \textcolor{BrickRed}{71.9}  & \textcolor{BrickRed}{88.9}  & \textcolor{RoyalBlue}{36.9}  & \textcolor{RoyalBlue}{17.3}  & \textcolor{BrickRed}{\textbf{49.1}} \\
\bottomrule
\end{tabular}
\caption{Comparison of LLM performance (mean@1 \& mean@32) across multiple text-only mathematical reasoning benchmarks. For each base model, the top-performing and runner-up results of RLVR algorithms within each column are marked in \textcolor{BrickRed}{red} and \textcolor{RoyalBlue}{blue}, respectively. $^{\dagger}$ indicates our reproduced results using publicly available models and standard evaluation code. “–” denotes missing results due to unavailable models.}
\label{tab:text-only-main-res}
\end{table*}

To establish the effectiveness of ReLaX, we perform extensive experiments on both VLMs and LLMs at the 3B and 7B scales. We begin in Section~\ref{sec:exp_settings} by outlining some important experimental setups. Section~\ref{sec:exp_res} then reports results on a broad suite of multimodal and text-only mathematical reasoning benchmarks, comparing ReLaX against state-of-the-art LRMs. Section~\ref{sec:exp_ablation} presents an ablation study illustrating the role of each component in ReLaX. Finally, beyond the benchmarking results, Section~\ref{sec:exp_compare} offers a comprehensive comparison with previous work that focus on token space.

\subsection{Experimental Settings}
\label{sec:exp_settings}
We conduct experiments on both VLMs and LLMs within the GRPO framework, which is widely adopted in recent studies on the emergence of LRMs. Below, we outline the key experimental configurations used in our experiments:

\subsubsection{Training Data and Benchmarks} 
VLMs are trained on the ViRL39K dataset from~\cite{wang2025vlrethinker}, which contains 38,870 curated and verifiable multimodal question–answer pairs. We adjust the query special tokens to align with our implementation framework. To comprehensively evaluate multimodal reasoning performance, $7$ challenging multimodal reasoning benchmarks are involved, including multidisciplinary datasets (MMMU~\cite{10656299}, MMStar~\cite{chen2024are}, EMMA~\cite{hao2025can}) and mathematical datasets (MathVista~\cite{lu2024mathvista}, MathVerse~\cite{10.1007/978-3-031-73242-3_10}, MathVision~\cite{wang2024measuring}, and DynaMath~\cite{zou2025dynamath}). We report the mean@1 accuracy using greedy decoding.

For LLMs training, we construct a merged dataset by combining the DAPO-Math-17K corpus~\cite{yu2025dapo} with the Level 3–5 subsets of the MATH training set~\cite{hendrycks2021measuring}, yielding approximately 22K mathematical queries with a broader range of difficulty. For evaluation, we use MATH500 (500 random samples from the MATH test set), Minerva~\cite{lewkowycz2022solving}, AMC 2022 \& 2023~\cite{ouyang2022training}, and AIME 2024 \& 2025~\cite{li2024numinamath}. We report mean@1 on MATH500 and Minerva, and mean@32 on AMC and AIME for robust evaluation due to the limited size of these test sets.

\subsubsection{Baselines and Implementations}
The multimodal baselines include MM-Eureka~\cite{meng2025mm}, Vision-R1~\cite{huang2025vision}, R1-VL~\cite{zhang2025r1}, OpenVLThinker~\cite{deng2025openvlthinker}, VL-Rethinker~\cite{wang2025vlrethinker}, and SRPO~\cite{jiang2023understanding}.
For text-only mathematical reasoning, we benchmark ReLaX against a set of recent RLVR algorithms, including SimpleRL~\cite{zeng2025simplerl}, DAPO~\cite{yang2025dcpo}, KL-Cov~\cite{cui2025entropy}, R1-zero-Div~\cite{yao2025diversity}, and FR3E~\cite{zheng2025first}. Our method is implemented on the VeRL codebase~\cite{sheng2024hybridflow}. Additional details on baselines, training hyperparameters, evaluation settings, and our ReLaX implementation are provided in the Supplementary Material.

\subsection{Main Results}
\label{sec:exp_res}

\subsubsection{Multimodal Reasoning} 
For multimodal experiments, we adopt Qwen2.5-VL-Instruct as the base model. As shown in Table~\ref{tab:multimodal_comparison}, ReLaX-VL-3B and ReLaX-VL-7B yield absolute improvements of $8.3$ and $5.3$ in average mean@1 accuracy across $7$ multimodal benchmarks, respectively, over their corresponding base models.
Notably, ReLaX-VL-7B achieves an average score of $53.2$, establishing a new state-of-the-art among existing 7B scale multimodal reasoning models and surpassing the previous best, VL-Rethinker-7B ($52.5$). At the 3B scale, ReLaX-VL-3B also shows strong competitiveness, outperforming several previous 7B-level models, including R1-VL ($40.9$) and OpenVLThinker ($45.3$). 

The training dynamics shown in Fig.~\ref{fig:ReLaX-training-dynamics} further highlight how ReLaX enhances the capability of multimodal LRMs. Under vanilla GRPO (gray), both policy entropy and DSD exhibit a rapid decline within the first $50$ steps. This indicates that the model quickly collapses into a rigid pattern—both in its internal latent dynamics and output tokens—ultimately leading to stagnant policy improvement due to reduced flexibility. In contrast, ReLaX maintains substantially more diverse latent dynamics and stabilizes the entropy at a higher yet well-regulated level. This balanced behavior enables the policy to continue improving and ultimately achieves a relative performance gain of $10\%$ on both the 3B and 7B models. These results demonstrate that ReLaX unlocks a substantial amount of previously untapped potential by explicitly facilitating the exploration–exploitation tradeoff during RLVR—an aspect that prior MLLM work has largely overlooked.

\subsubsection{Text-only Mathematical Reasoning} 
For text-only reasoning, we conduct experiments on three base models: Qwen2.5-3B-Base, Qwen2.5-7B-Base, and Qwen2.5-7B-Math. We compare ReLaX against existing RLVR algorithms, including the GRPO baseline (referred to as SimpleRL~\cite{zeng2025simplerl} for 7B-scale models) and several variants designed to promote the policy exploration at token space. As shown in Table~\ref{tab:text-only-main-res}, ReLaX achieves substantial improvements over all three base models as well as their vanilla GRPO counterparts. Moreover, ReLaX outperforms existing GRPO variants across $6$ mathematical reasoning benchmarks. In particular, ReLaX surpasses the previous state-of-the-art publicly available baseline, FR3E~\cite{zheng2025first}, by $4.3$ on Qwen2.5-7B-Base and $6.3$ on Qwen2.5-7B-Math. 
The training dynamics in text-only experiments on Qwen2.5-Base with 3B and 7B scales can be found in Supplementary Material. 
To showcase that ReLaX generalizes across model families, we additionally evaluate it on Llama3.2-3B-Instruct and Qwen3-4B. The corresponding results are provided in the Supplementary Material.

\subsection{Ablation Study}
\label{sec:exp_ablation}
We perform ablation studies to analyze the DSD-based regularization in ReLaX from two perspectives.

\begin{figure}[htbp]
  \centering
  \begin{subfigure}{0.23\textwidth}
    \centering
    \includegraphics[width=\linewidth]{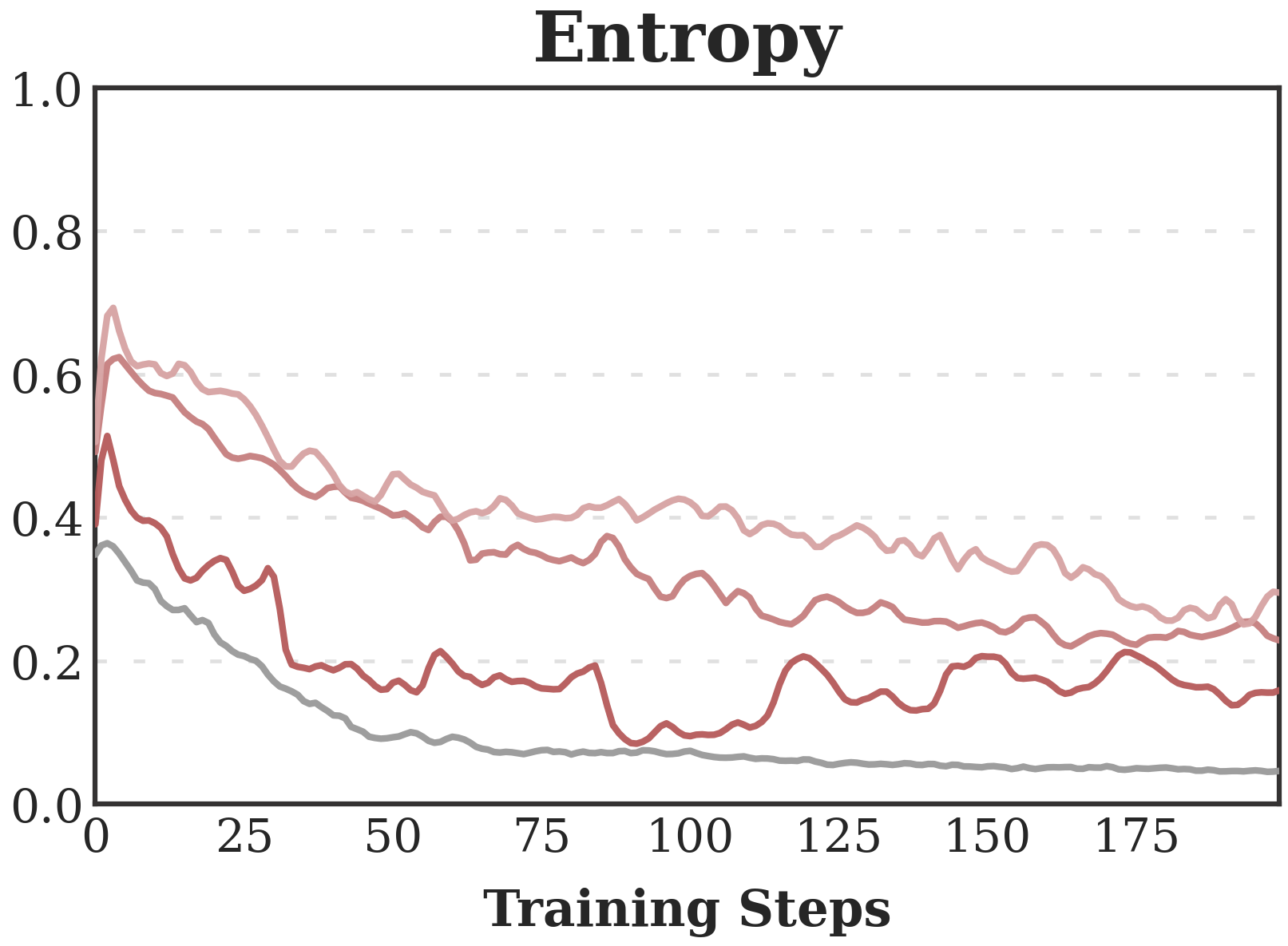}
    \caption{}
    \label{fig:mllm-ablation1-a}
  \end{subfigure}
  \begin{subfigure}{0.23\textwidth}
    \centering
    \includegraphics[width=\linewidth]{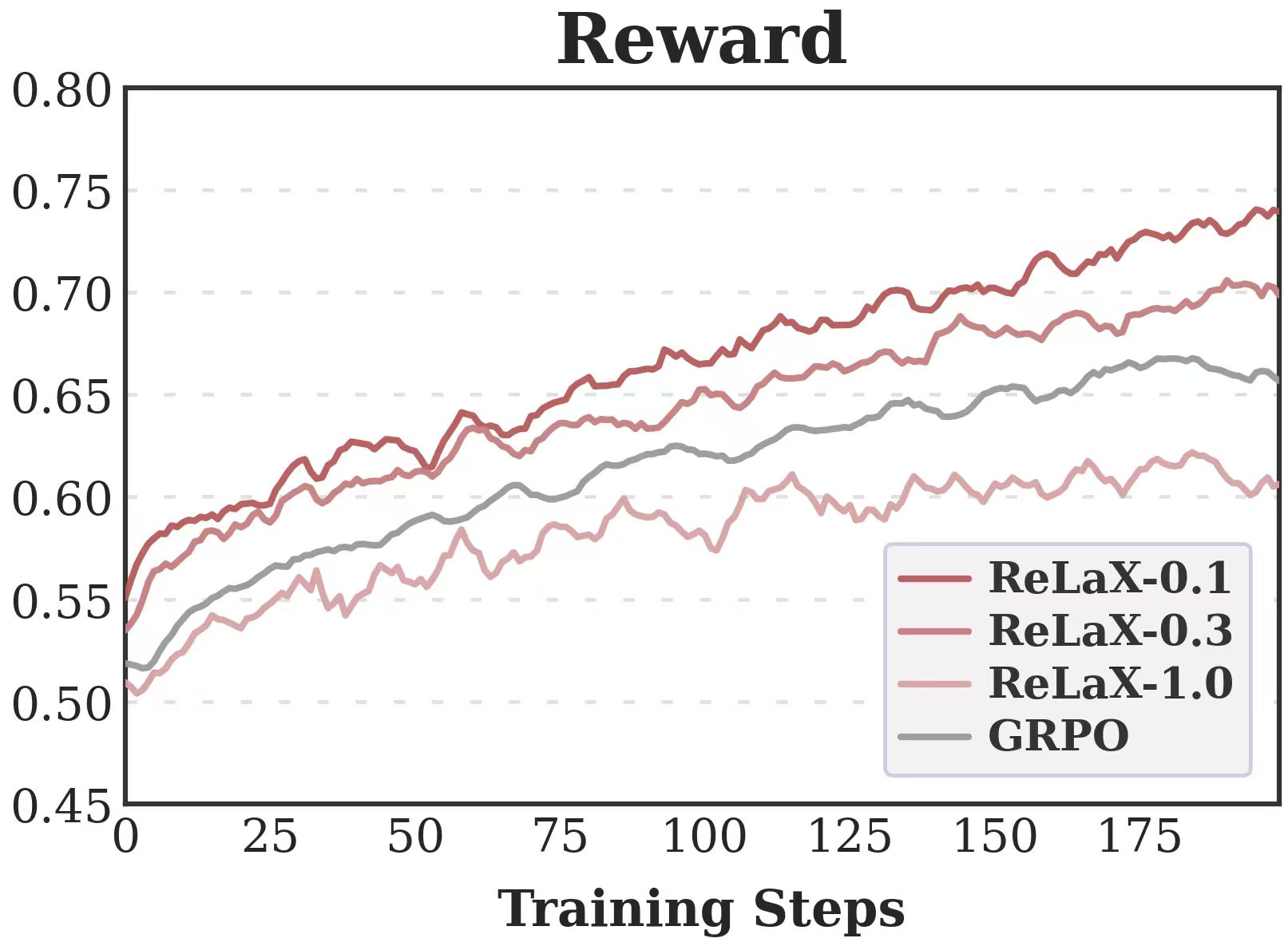}
    \caption{}
    \label{fig:mllm-ablation1-b}
  \end{subfigure}
  \caption{Training dynamics of policy entropy and reward on 3B scale Qwen2.5-VL-3B models by ReLaX with different DSD-based regularization coefficients (1.0, 0.3, 0.1, 0).}
  \label{fig:ablation-1}
\end{figure}

\noindent \textbf{Hyperparameters Sensitivity.} We first evaluate the impact of the DSD-based regularization coefficient $\alpha$, which governs the strength of encouraging the policy to explore more diverse latent dynamics. As shown in Fig.~\ref{fig:ablation-1}, increasing this coefficient leads to a pronounced rise in policy entropy, indicating that nudging the model away from rigid internal computation regimes effectively mitigates entropy collapse. However, higher entropy does not necessarily translate into better policy convergence. We observe that ReLaX achieves the highest policy reward when the coefficient is set to 0.1. A coefficient of 0.3 yields only a marginal improvement over vanilla GRPO, while an overly strong coefficient (e.g., 1.0) harms performance. Additional results on Koopman dimension $m$ and DSD threshold $\xi$ are provided in the Supplementary Material. 

\begin{figure}
    \centering
    \includegraphics[width=\linewidth]{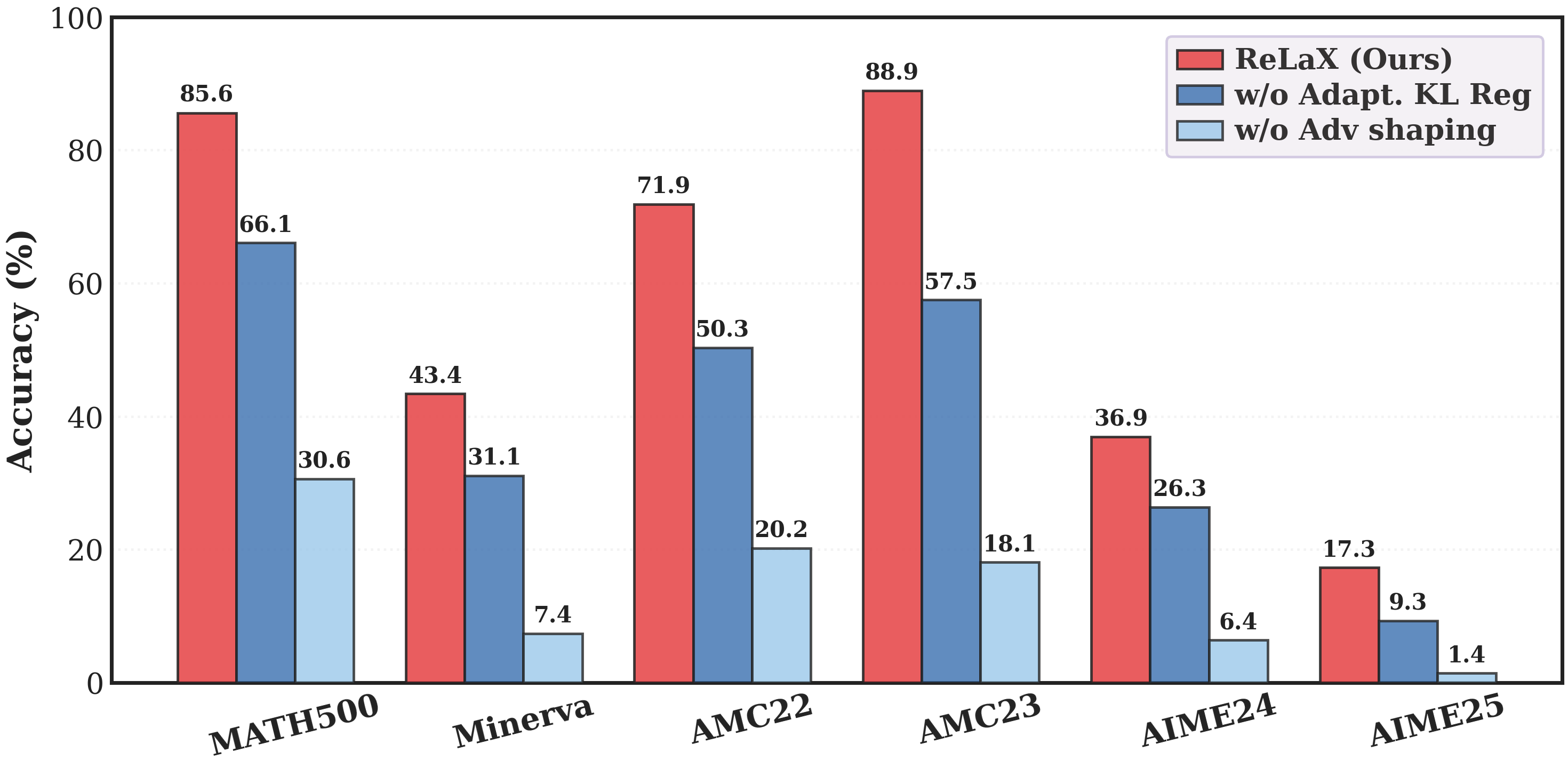}
    \caption{Evaluation results from the ablation study on Qwen2.5-7B-Math for text-only reasoning tasks. Results by full ReLaX is highlighted in red, while the dark-blue and light-blue bars respectively correspond to its ablations without adaptive KL regularization and advantage shaping. }
    \label{fig:ablation-2}
\end{figure}

\noindent \textbf{Adaptive KL Regularization \& Advantage Shaping.}
To stabilize policy optimization, ReLaX incorporates adaptive KL regularization and advantage shaping. We evaluate their contributions via ablations on Qwen2.5-7B-Math across six mathematical reasoning benchmarks. As shown in Fig.~\ref{fig:ablation-2}, removing adaptive KL regularization applies uniform penalties across trajectories regardless of their DSD scores, leading to consistently degraded performance, consistent with prior findings~\cite{yu2025dapo, cui2025entropy}. In contrast, ReLaX penalizes only trajectories with overly dispersed dynamics. Removing advantage shaping causes a more severe drop: exploration becomes indiscriminate in latent space, including directions with negative or uninformative rewards, resulting in training collapse and even worse performance than the base model. 

Overall, these results suggest that ReLaX achieves a balance between exploration and exploitation through a principled choice of hyperparameters and beneficial trajectory selection, enabling controlled, conditional latent exploration while maintaining stable learning.

\subsection{Comparative Analysis}
\label{sec:exp_compare}

ReLaX operates beyond token-level manipulation by reshaping intrinsic latent reasoning dynamics. While it demonstrates clear performance gains on text-only reasoning benchmarks (Table~\ref{tab:text-only-main-res}), its advantages over token-level methods in multimodal reasoning, as well as its impact on specific reasoning behaviors, remain unclear. We therefore compare it with representative token-level methods in this section through multimodal generalization and qualitative analysis.

\subsubsection{Multimodal Generalization}
\begin{figure}[t]
  \centering
  \begin{subfigure}{0.47\textwidth}
    \centering
    \includegraphics[width=\linewidth]{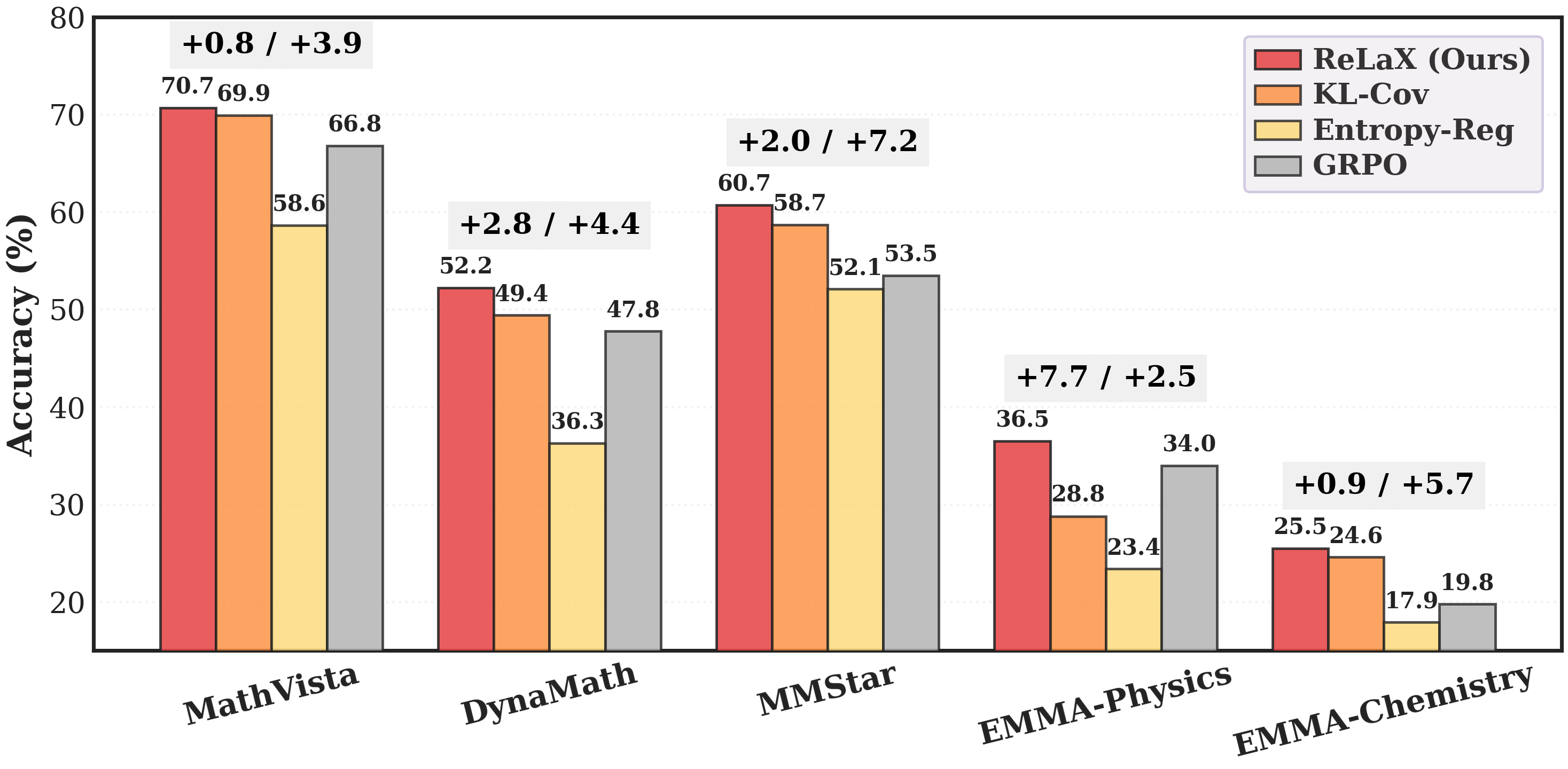}
    \caption{}
    \label{fig:mllm-compare-a}
  \end{subfigure}
  \begin{subfigure}{0.23\textwidth}
    \centering
    \includegraphics[width=\linewidth]{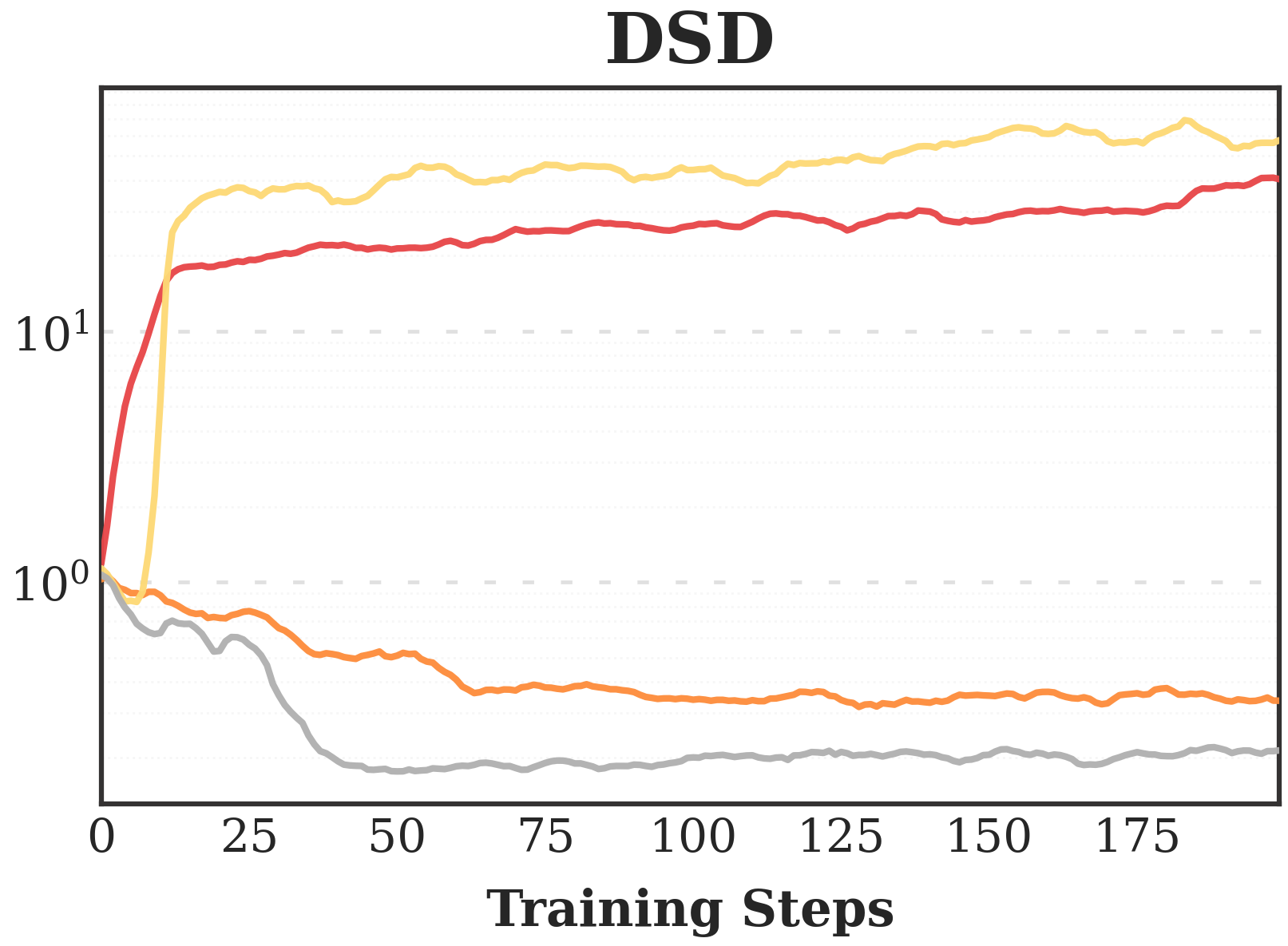}
    \caption{}
    \label{fig:mllm-compare-b}
  \end{subfigure}
  \begin{subfigure}{0.23\textwidth}
    \centering
    \includegraphics[width=\linewidth]{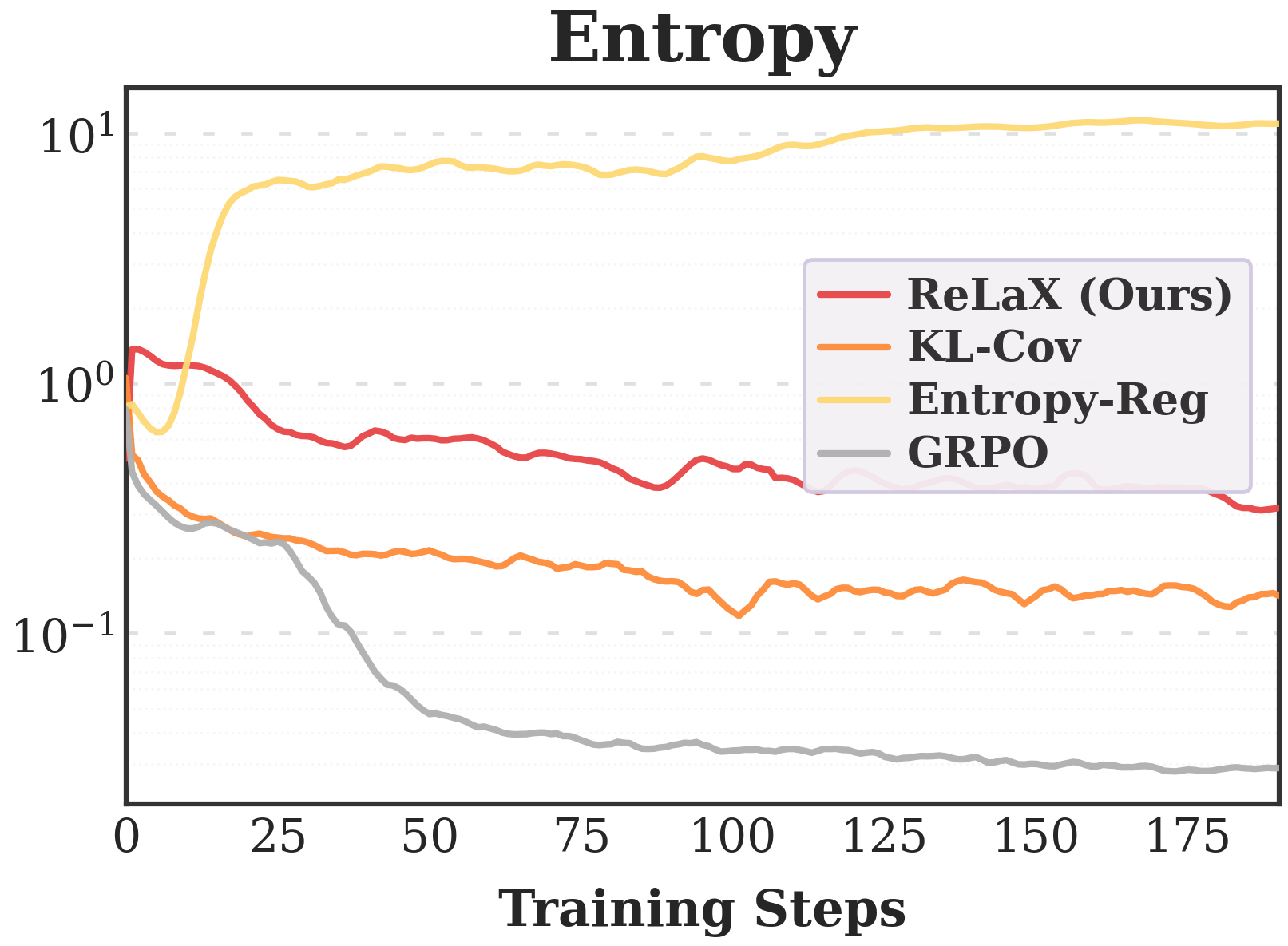}
    \caption{}
    \label{fig:mllm-compare-c}
  \end{subfigure}
  \caption{Comparison of RLVR training methods on Qwen2.5-3B-VL. (a) Evaluation results across 5 multimodal reasoning benchmarks. The values above each bar denote ReLaX’s performance gains over KL-Cov (left) and vanilla GRPO (right). (b) and (c) Training dynamics of policy DSD and policy entropy, respectively.}
  \label{fig:mllm-compare-3b}
\end{figure}

We evaluate ReLaX against entropy regularization~\cite{yao2025diversity} and KL-Cov~\cite{cui2025entropy} on Qwen2.5-VL-3B. As shown in Fig.~\ref{fig:mllm-compare-3b}, ReLaX consistently outperforms both methods across multimodal benchmarks. Naively extending entropy regularization to multimodal training increases both DSD and token-level entropy, but at the cost of semantic drift that degrade performance. While KL-Cov provides modest gains over vanilla GRPO, it does not prevent the rigidification of latent dynamics, suggesting that promoting token-level exploration have limited influence on the latent space where cross-modal computation takes place. Interestingly, KL-Cov performs relatively better on multimodal mathematics benchmarks such as MathVista, MathVerse, and MathVision (full results in the Supplementary Material), where visual grounding is less critical. In contrast, ReLaX achieves substantially larger gains on more visually grounded tasks (e.g., +7.7 on EMMA-Physics). 

These findings indicate that explicitly encouraging flexible latent reasoning dynamics is more appropriate for MLLMs, as it facilitates a more effective exploration–exploitation trade-off and leads to better generalization in multimodal reasoning.

\subsubsection{Qualitative Results}

We further present qualitative case studies to examine reasoning behaviors underlying the benchmark results. We first analyze a mathematic query from AMC23~\cite{ouyang2022training}, comparing the models trained with ReLaX and entropy regularization (R1-zero-Div~\cite{yao2025diversity}) from Qwen2.5-7B-Math. As shown in Supplymentary Material Tab.~\ref{tab:case-study-amc}, both models produce the correct final answer and attempt \emph{self-verification}. However, ReLaX-7B performs meaningful validation by invoking relevant mathematical principles, whereas R1-zero-Div generates Python code for verification—an ineffective strategy given the lack of code execution. This behavior suggests that simply encouraging uncertainty on token generation may lead to hallucinated or inefficient reasoning. Additionally for VLMs, We compare ReLaX with KL-Cov on the Qwen2.5-VL-3B model using a sample from the DynaMath dataset~\cite{zou2025dynamath}, which evaluates reasoning robustness by testing whether models remain consistent across variants of the same question (e.g., modified numerical values or visual details). As shown in Supplementary Material Tab.~\ref{tab:case-study-dynamath}, ReLaX demonstrates consistent visual understanding across all variants, yielding correct answers. In contrast, KL-Cov fails on several variants due to misinterpretation of visual cues or incorrect reasoning, indicating degraded visual comprehension and reasoning capability. An additional case of this perspective from MMMU~\cite{10656299} is included in the Supplementary Material Tab.~\ref{tab:case-study-MMMU-Physics}.

%% file: sec/5_conclusion.tex
\section{Conclusion}
\label{sec:conclusion}
In this work, we introduced ReLaX, a new RLVR training paradigm that explicitly promotes heterogeneity in latent dynamics during policy optimization. By facilitating more flexible internal computations for LRMs, ReLaX effectively mitigates the performance saturation induced by mode collapse in RLVR. Extensive evaluations across multiple benchmarks show that ReLaX consistently improves reasoning capability under RLVR paradigm across diverse model families, outperforming existing methods that focus on token-level diversity. These results indicate that steering exploration via latent dynamics offers a more principled solution to the exploration–exploitation trade-off, and point to latent space as a powerful and scalable frontier for advancing LRM capabilities.


%% file: sec/X_suppl.tex
\clearpage
\setcounter{page}{1}
\setcounter{section}{0}
\maketitlesupplementary

\begin{strip}
\vspace{-25pt}
\begin{tcolorbox}[
    colback=blue!5!white, 
    colframe=blue!50!black, 
    title=\textbf{Overview},
    arc=2mm, 
    boxrule=0.5pt
]
This Supplementary Material provides more details and results of the proposed ReLaX.

\begin{itemize}
    \item In Supp. Sec. 1, we discuss the \textbf{related work} of recent GRPO variants that aim to address the ineffective exploration through the lens of token-level entropy.
    \item In Supp. Sec. 2, we provide supplementary background on \textbf{Koopman dictionary learning}, along with t-SNE visualizations of latent trajectories under the learned dictionary and the \textbf{pseudocode} of ReLaX.
    \item In Supp. Sec. 3, We comprehensively describe our \textbf{experimental setup}.
    \item In Supp. Sec. 4, \textbf{additional experimental results} are provided to complement the analyses in Sec.~\ref{sec:exp}.
\end{itemize}
\end{tcolorbox}
\end{strip}

\section{Related Work}
\label{suppl-sec:related-work}

Although RLVR has achieved impressive gains in mathematical, coding, and visual reasoning tasks, recent studies~\cite{yue2025does,wu2025invisible} suggest that its improvements largely arise from enhanced sampling efficiency rather than true capability gains. What is worse, as the model becomes increasingly confident and generates more convergent outputs, policy learning suffers from reduced exploration, leading to saturation and a performance bottleneck—a central challenge in balancing exploration and exploitation~\cite{cui2025entropy,shen2025entropy}.
To mitigate this issue, recent research focuses on maintaining sufficient entropy in the model’s action space to promote effective exploration. 

Early methods~\cite{yao2025diversity,lei2025revisiting} seek to maximize policy entropy directly or incorporate it into the reward signal~\cite{cheng2025reasoning}. Such methods, however, require careful tuning of weighting coefficients: insufficient weighting fails to prevent collapse, while excessive weighting may induce semantic drift. 

To regulate policy entropy in a more principled and controllable way, some research focuses on identifying structural factors that drive changes in entropy. DAPO~\cite{yu2025dapo} observes that the standard PPO/GRPO clipping range, though stabilizing the training, imposes an overly restrictive upper bound that suppresses exploration. Relaxing this bound via the clip-higher strategy increases credit for positively rewarded tokens and promotes broader exploration. Building on this insight, DCPO~\cite{yang2025dcpo} introduces a dynamic, adaptive clipping mechanism that affords finer control over update magnitudes. Cui et al.~\cite{cui2025entropy} further reveal that the covariance between action probabilities and logit variations strongly predicts entropy reduction; accordingly, they propose KL-Cov and Clip-Cov to selectively penalize or clip high-covariance tokens most likely to induce entropy collapse.

Another line of approaches focuses on allocating policy-update credit selectively to high-entropy tokens. For instance, ~\cite{wang2025beyond} updates only the top 20\% high-entropy forking tokens and reports substantial performance gains. FR3E~\cite{zheng2025first} extends this idea by identifying high-entropy tokens during sampling and expanding rollouts specifically along these positions; subsequent policy optimization is then performed on these augmented trajectories, effectively enhancing exploration. CURE~\cite{li2025cure} further generalizes FR3E by stochastically selecting high-entropy tokens to mitigate selection bias. To consolidate the improved exploration into stronger exploitation, CURE additionally applies DAPO~\cite{yu2025dapo} in a second training stage.


\section{Details on Koopman Dictionary Learning}
\label{suppl-sec:koopman}
This section provides background and details the procedure for learning the Koopman dictionary, which maps the model’s nonlinear, high-dimensional hidden states into a linear, tractable representation. An illustrative example The pseudocode for dictionary learning and the complete ReLaX training workflow is given in Algorithm~\ref{alg:relax}.

\subsection{Residual DMD (ResDMD)}
Approximating the Koopman operator requires projecting its infinite-dimensional dynamics onto a finite set of observables~\cite{li2017extended}. This projection inevitably introduces spurious eigenvalues, a problem exacerbated for large reasoning models whose hidden states are extremely high-dimensional (e.g., 3584 for Qwen2.5-7B) and evolve over long contexts. In such settings, standard DMD approaches overfit transient fluctuations, suffer numerical instability, and produce spectral artifacts that obscure the true temporal modes.

ResDMD~\cite{colbrook2024rigorous} improves spectral fidelity by diagnosing and filtering corrupted eigenvalues using a residual test. Given snapshot matrices $\mathcal{V}$ and $\mathcal{V}^+$ and an approximate Koopman operator~$\mathcal{K}$, the squared residual of an eigenpair $(\lambda, v)$ is estimated as:
\begin{equation}
\label{eq:resdmd}
\begin{aligned}
\mathrm{res}(\lambda, v)^2 :=\;
v^* \big[ & (\mathcal{V}^+)^{*}\mathcal{V}^+
 - \lambda (\mathcal{V}^{*}\mathcal{V}^+)^{*} 
 - \bar{\lambda}\, \mathcal{V}^{*}\mathcal{V}^+  \\
& \quad +\, |\lambda|^{2}\, \mathcal{V}^{*}\mathcal{V} \big] v.
\end{aligned}
\end{equation}
This residual measures how well the eigenpair satisfies the least-squares formulation and, therefore, how faithfully it captures the underlying dynamics. Eigenpairs with large residuals are classified as spurious and removed. A limitation, however, is that this diagnostic only becomes available once the Koopman spectrum is computed with a fixed set of observables. Since choosing an effective dictionary is itself a central and challenging problem, a natural question arises: can we learn the dictionary jointly while minimizing spectral residuals?

\subsection{ResKoopNet}
ResKoopNet~\cite{xu2025reskoopnet} addresses this by parameterizing the Koopman dictionary with an MLP and optimizing it directly through the residual objective in Eq.~\ref{eq:resdmd}. The resulting residual loss is given in Eq.~\ref{eq:reskoopnet}, with the full derivation provided in the original paper.

In our implementation, the dictionary is parametrized by a single feedforward layer with a Sigmoid activation. It is trained on the hidden states collected during the first step of sampling, yielding a general set of observables that defines a globally shared linear representation space for the subsequent policy optimization. The complete procedure is provided in Algorithm~\ref{alg:relax}.

\begin{figure}[!ht]
  \centering
  \includegraphics[width=0.99\linewidth]{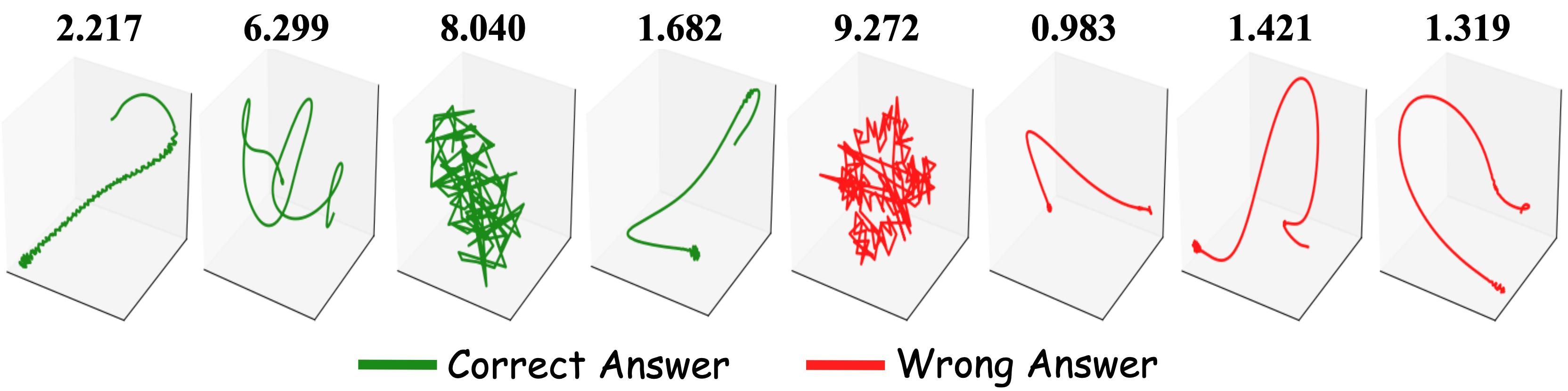}
  \caption{t-SNE of latent trajectories (DeepSeek-Math-7B). While sample-wise DSD does not strongly correlate with correctness, policy optimization benefits from consistently flexible internal computations, as reflected by higher DSD.}
  \label{fig:traj_dsd}
\end{figure}

\RestyleAlgo{ruled}
\begin{algorithm*}[!hbt]
\caption{ReLaX}
\label{alg:relax}
\DontPrintSemicolon
\SetAlgoLined

\SetKwInput{KwInput}{Input}
\SetKwInput{KwOutput}{Output}
\SetKwFunction{FitKoopman}{FitKoopmanDict}
\SetKwProg{Fn}{Function}{:}{}

\KwInput{Dataset $\mathcal{D}$, policy $\pi_\theta$, reference policy $\pi_{\text{ref}}$, 
batch size $B$, group size $R$, learning rate $\eta$, total training steps $\mathcal{S}$}
\KwOutput{Optimized policy $\pi_\theta$}

\BlankLine
Set iteration counter $s \leftarrow 0$\;

\While{$s \leq \mathcal{S}$}{

    Sample a batch of queries $Q \sim \mathcal{D}$\;
    Initialize gradient accumulator $\nabla \mathcal{J} \leftarrow 0$\;

    \tcc{1. Sampling}
    \ForEach{$q \in Q$}{
        \tcc{vLLM inference}
        Generate $R$ completions $O = \{o^1,\dots,o^R\}$ from $\pi_{\theta_{\text{old}}}(\cdot \mid q)$\;
        Calculate $\mathrm{reward}(Q, O)$ for completions\;
        \tcc{Transformers inference}
        Compute log probabilities\;
        Collect hidden states $X = \{x^1,\dots,x^R\}$ from the final hidden layer\;
    }

    \BlankLine
    \tcc{2. Koopman Dictionary Learning (First step only)}
    \If{$s = 0$}{
        Form batch-level hidden-state set: $X_B = \bigcup_{q \in Q} X^q$\;
        $g \leftarrow \FitKoopman(X_B)$\;
    }

    \BlankLine
    \tcc{3. Policy Optimization}
    \ForEach{$x \in X$}{
        Compute $\mathrm{DSD}(x)$ using Eq.~\ref{eq:dsd}\;
    }

    Compute ReLaX objective $\mathcal{J}(\theta)$ using Eq.~\ref{eq:relax_obj}\;
    Update policy: $\theta \leftarrow \theta + \eta \nabla_\theta \mathcal{J}(\theta)$\;

    \BlankLine
    $s \leftarrow s + 1$\;
}

\BlankLine

\Fn{\FitKoopman($X$)}{
    \KwInput{Hidden states $X \in \mathbb{R}^{BR \times T \times d}$}

    Select batch size $B_g$, total optimization steps $\mathcal{S}_g$, Koopman dimension $m$, and learning rate $\eta_g$\;
    Initialize dictionary network $g$ with parameters $W \in \mathbb{R}^{d \times m}$ and sigmoid activation $\sigma$\;

    Set iteration counter $s \leftarrow 0$\;

    \While{$s \leq \mathcal{S}_g$}{
        Sample hidden state batch $X_B \sim X$\;
        Construct consecutive temporal snapshots $\mathcal{V}_B, \mathcal{V}_B^{+} \in \mathbb{R}^{B_g \times (T-1) \times d}$\;
        Estimate Koopman operator $\mathcal{K}$ using Eq.~\ref{eq:koop-approx}\;
        Compute residual loss $\mathcal{J}_g(W)$ using Eq.~\ref{eq:reskoopnet}\;
        Update dictionary parameters $W \leftarrow W - \eta_g \nabla_W \mathcal{J}_g(W)$\;

        $s \leftarrow s + 1$\;
    }

    \KwRet{Learned Koopman dictionary $g$}\;
}

\end{algorithm*}

\section{Detailed Experimental Settings}
\label{suppl-sec:exp-setting}
Our experiments were run on multiple GPU clusters, each equipped with 8× NVIDIA A100 (80GB). To improve training efficiency, we enabled actor–learner collocation via VeRL~\cite{sheng2024hybridflow}. The primary hyperparameters on finetuning both LLMs and VLMs, including those for the actor, trainer, and ReLaX-specific settings, are summarized in Table \ref{tab:rl-params}. All experiments employ identical configurations for optimizing the Koopman dictionary, using the Adam optimizer with a learning rate of $10^{-4}$ and a batch size of 64.

\begin{table}[h]
\centering
\small
\setlength{\tabcolsep}{6pt}
\renewcommand{\arraystretch}{1.15}
\begin{tabular}{
>{\raggedright\arraybackslash}p{0.52\linewidth}
>{\centering\arraybackslash}p{0.38\linewidth}
}
\toprule
\textbf{Parameter} & \textbf{Value} \\
\midrule
\multicolumn{2}{c}{\textbf{Actor}} \\
\midrule
Maximum response length       & 3072 \\
Temperature                   & 1.0 \\
top p                         & 1.0 \\
top k                         & -1 \\
Number of rollouts per prompt  & 16 \\
\midrule
\multicolumn{2}{c}{\textbf{Trainer}} \\
\midrule
Batch size                              & 512 \\
Generate size for sampling              & 2048 \\
Optimizer                               & AdamW \\
Adam betas                              & (0.9, 0.95) \\
Gradient norm clipping                  & 1.0 \\
Learning rate scheduler                 & Constant \\
Learning rate                           & $10^{-6}$ \\
\midrule
\multicolumn{2}{c}{\textbf{ReLaX-Specific}} \\
\midrule
Exploration coefficient $\alpha$        & 0.1 \\
KL loss coefficient $\beta$             & 0.01 \\
DSD threshold $\xi$                           & 10 \\
Koopman operator dimension $m$              & 50 \\
\bottomrule
\end{tabular}
\caption{Hyperparameters for RLVR training used in our experiments. The same settings are applied to both VLM and LLM experiments.}
\label{tab:rl-params}
\end{table}

The benchmarking results reported in Tables \ref{tab:multimodal_comparison} and \ref{tab:text-only-main-res} are compiled from published papers and our evaluations on publicly available checkpoints using open-source toolkits. Specifically, VLMs are evaluated with VLMEvalKit \footnote{https://github.com/open-compass/VLMEvalKit}, and LLMs are evaluated following Qwen2.5-Math \footnote{https://github.com/QwenLM/Qwen2.5-Math}. The evaluation hyperparameters for both VLMs and LLMs are listed in Table \ref{tab:eval-params}.

\begin{table}[h]
\centering
\small
\setlength{\tabcolsep}{6pt}
\renewcommand{\arraystretch}{1.15}
\begin{tabular}{
>{\raggedright\arraybackslash}p{0.52\linewidth}
>{\centering\arraybackslash}p{0.38\linewidth}
}
\toprule
\textbf{Parameter} & \textbf{Value} \\
\midrule
\multicolumn{2}{c}{\textbf{VLM evaluation}} \\
\midrule
top p      & 0.1 \\
Maximum response length             & 10240 \\
Temperature              & 0 \\
LLM-as-Judge             & GPT-4o-mini  \\
\midrule
\multicolumn{2}{c}{\textbf{LLM evaluation}} \\
\midrule
top p      & 0.1 \\
Maximum response length             & 10240 \\
Temperature (mean@1)              & 0 \\
Temperature (mean@32)              & 1 \\
\bottomrule
\end{tabular}
\caption{Hyperparameters used for Evaluations.}
\label{tab:eval-params}
\end{table}

Finally, the chat templates employed in VeRL for model training are presented below. During evaluation, we use the evaluation codebase’s default system prompt to ensure fair and comparable results.

\begin{AIbox}{Chat template for Visual-Language training.}
    System: Given the images and the question, please reason step by step, and put your final answer within \textbackslash boxed\{\}.\\
    User: $\langle$Question$\rangle$\\
    Assistant: 
\end{AIbox}

\begin{AIbox}{Chat template for text-only training.}
    System: Please reason step by step, and put your final answer within \textbackslash boxed\{\}.\\
    User: $\langle$Question$\rangle$\\
    Assistant: 
\end{AIbox}

\begin{figure*}
    \centering
    \includegraphics[width=.95\linewidth]{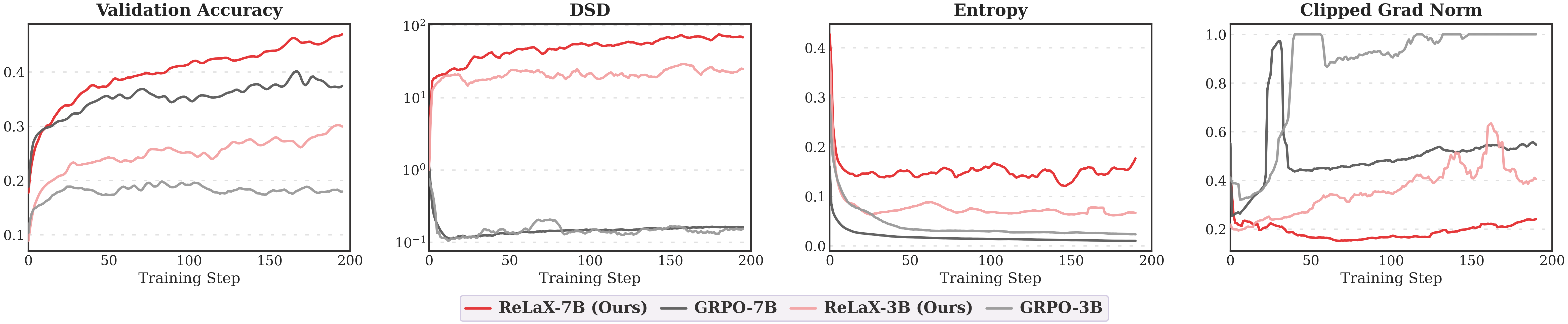}
    \caption{Comparison of training dynamics for \textit{Validation Accuracy}, \textit{DSD}, \textit{Entropy}, and \textit{Clipped Gradient Norm} under ReLaX (red) and vanilla GRPO (gray) on Qwen2.5-Base at the 3B and 7B scales.}
    \label{fig:ReLaX-training-dynamics-Lan}
\end{figure*}

\begin{table*}[ht]
\centering
\small
\setlength{\tabcolsep}{5pt}
\renewcommand{\arraystretch}{1.15}
\begin{tabular}{lcccccccccc}
\toprule
\multirow{2}{*}{\textbf{Model}} & 
\multirow{2}{*}{\textbf{Size}} & 
\multicolumn{1}{c}{\textbf{MATH500}} & 
\multicolumn{1}{c}{\textbf{Minerva}} & 
\multicolumn{1}{c}{\textbf{AMC22}} & 
\multicolumn{1}{c}{\textbf{AMC23}} & 
\multicolumn{1}{c}{\textbf{AIME24}} & 
\multicolumn{1}{c}{\textbf{AIME25}} & 
\multirow{2}{*}{\textbf{Average}} \\
\cmidrule(lr){3-8}
 &  & \textit{Mean@1} & \textit{Mean@1} & \textit{Mean@32} & \textit{Mean@32} & \textit{Mean@32} & \textit{Mean@32} &  \\
\midrule
Llama3.2-Instruct & 3B & 46.8 & 15.4 & 16.1$^{\dagger}$ & 22.0 & 8.5 & 0 & 18.1 \\
\quad +Vanilla GRPO\cite{shao2024deepseekmath} & 3B & 55.4 & 22.8 & 26.4 & 40.0 & 16.3 & 1.4 & 27.1 \\
\rowcolor{blue!8}
\quad \textbf{\textcolor{black}{+ReLaX (Ours)}} & 3B & 57.0 & 23.5 & 39.0 & 52.8 & 18.9 & 3.3 & 32.4 \\
\rowcolor{blue!8}
\quad \textbf{\textcolor{black}{$\Delta$ (ReLaX - GRPO)}} & 3B & \textcolor{BrickRed}{+1.6} & \textcolor{BrickRed}{+0.7} & \textcolor{BrickRed}{+12.6} & \textcolor{BrickRed}{+12.8} & \textcolor{BrickRed}{+2.6} & \textcolor{BrickRed}{+1.9} & \textcolor{BrickRed}{+5.3} \\
\midrule
Qwen3-Base & 4B & 63.8 & 28.3 & 29.1$^{\dagger}$ & 38.9 & 9.4 & 5.3 & 29.1 \\
\quad +Vanilla GRPO\cite{shao2024deepseekmath} & 4B & 83.0 & 38.9 & 42.6 & 51.2 & 24.9 & 23.8 & 44.1 \\
\quad +HICRA~\cite{wang2025emergent} & 4B & 89.0 & 42.5 & - & 54.0 & 31.0 & 27.6 & - \\
\rowcolor{blue!8}
\quad \textbf{\textcolor{black}{+ReLaX (Ours)}} & 4B & 90.2  & 48.5  & 52.6 & 64.5  & 30.9  & 27.6  & 52.3  \\
\rowcolor{blue!8}
\quad \textbf{\textcolor{black}{$\Delta$ (ReLaX - GRPO)}} & 4B & \textcolor{BrickRed}{+6.2} & \textcolor{BrickRed}{+9.6} & \textcolor{BrickRed}{+10.0} & \textcolor{BrickRed}{+13.3} & \textcolor{BrickRed}{+6.0} & \textcolor{BrickRed}{+3.8} & \textcolor{BrickRed}{+8.2} \\
\bottomrule
\end{tabular}
\caption{Supplemented comparison of LLM performance (mean@1 \& mean@32) trained from Llama3.2-Instruct and Qwen3 across multiple text-only mathematical reasoning benchmarks. The performance gains of ReLaX over the GRPO baseline are highlighted in \textcolor{BrickRed}{red}. $^{\dagger}$ indicates our reproduced results using publicly available models and standard evaluation code. “–” denotes missing results due to unavailable models.}
\label{tab:text-only-llama-qwen3}
\end{table*}

\begin{figure*}[t]
    \centering
    \includegraphics[width=\linewidth]{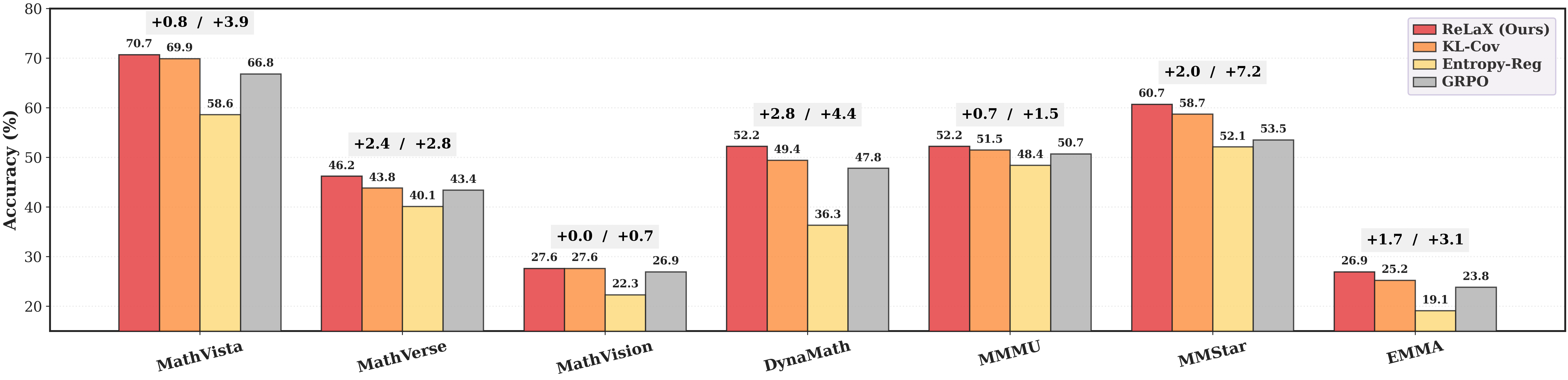}
    \caption{Complete comparison of multimodal benchmark performance for different training methods on Qwen2.5-3B-VL. This figure provides the extended results referenced in Fig.~\ref{fig:mllm-compare-a}. The values above the bars denote ReLaX’s performance gains over KL-Cov (left) and vanilla GRPO (right).}
    \label{fig:mllm-compare-benchmark-supp}
\end{figure*}

\begin{figure*}[t]
    \centering
    \includegraphics[width=\linewidth]{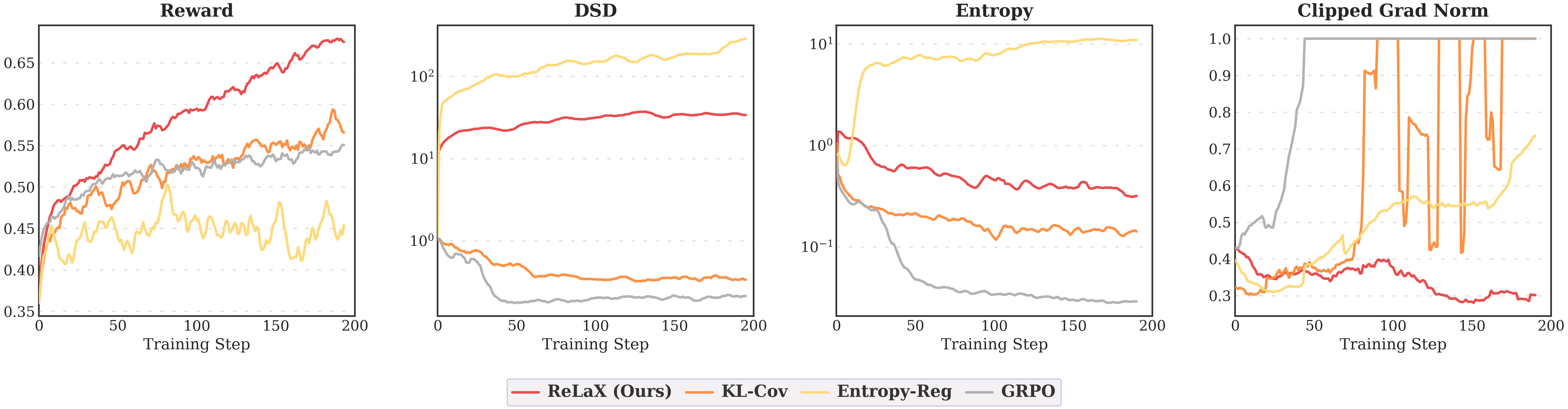}
    \caption{More training dynamics of vanilla GRPO, KL-Cov, Entropy Reg and our ReLaX on Qwen2.5-3B-VL. This figure provides the extended results for Fig.~\ref{fig:mllm-compare-b} and ~\ref{fig:mllm-compare-c}.}
    \label{fig:mllm-compare-dynamics-supp}
\end{figure*}

\section{Additional Results}
\label{suppl-sec:add-res}
In this section, we present the additional results and details to clarify some of the analytical experiments and substantiate our claims.
\subsection{Training Dynamics of Text-only Experiments}

We observe training dynamics (Fig.~\ref{fig:ReLaX-training-dynamics-Lan}) in the text-only scenario that mirror the phenomenon seen previously in the vision-language experiments. Compared with vanilla GRPO, ReLaX maintains higher policy entropy, more diverse latent dynamics, and longer response lengths. Moreover, ReLaX exhibits strong training stability, as evidenced by its consistently low gradient clipping rate. Together, these properties enable ReLaX to achieve substantial improvements in validation accuracy.

\subsection{Supplemented Experiments on Other Models}
Although models from the Qwen2.5 family are commonly used as base models for RLVR training in recent work, there has been an ongoing community discussion regarding potential evaluation set leaks.
To this end, we further include experiments on Llama3.2-3B-Instruct and Qwen3-4B-Base to demonstrate that ReLaX generalizes beyond specific foundation models. 

As shown in Table~\ref{tab:text-only-llama-qwen3}, our method consistently outperforms the GRPO baseline across both types of base models. To further compare with token-level methods, we additionally include results from HICRA~\cite{wang2025emergent}. ReLaX achieves substantial gains on Minerva and AMC, while remaining competitive on MATH500 and AIME. These results demonstrate that ReLaX generalizes effectively across different base models and maintains a stable advantage over token-level methods.

\subsection{Additional Results on Sensitivity Analysis}

\begin{figure}[t]
  \centering
  \includegraphics[width=0.99\linewidth]{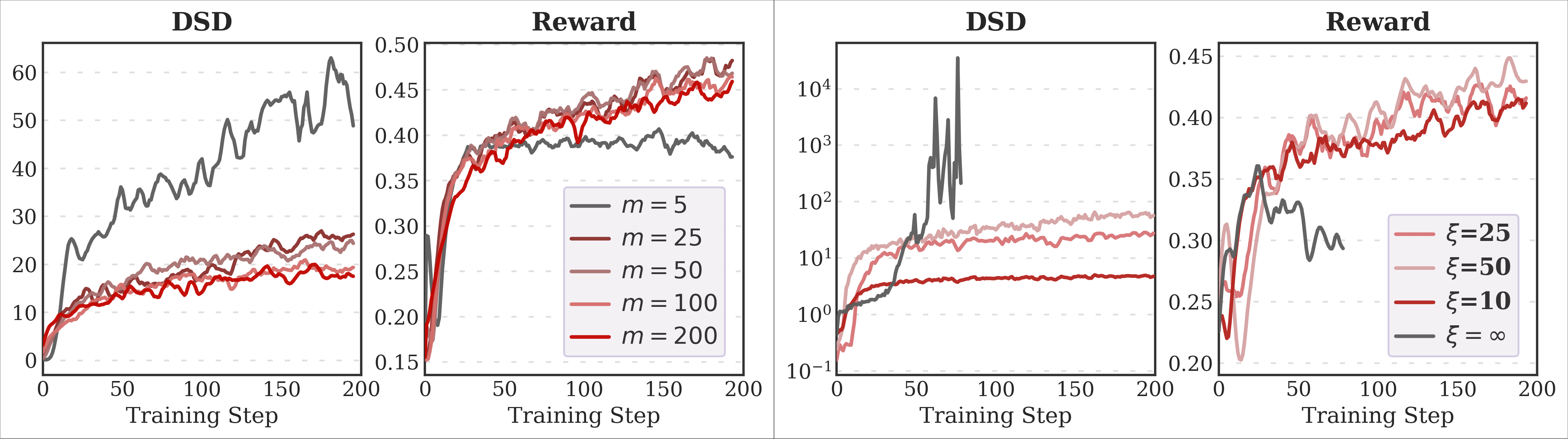}
    \vspace{-10pt}
   \caption{Sensitivity analysis for Koopman operator dimension $m$ and DSD threshold $\xi$ on Qwen2.5-3B.}
   \label{suppl-fig:sen-ana}
   \vspace{-20pt}
\end{figure}

Fig.~\ref{suppl-fig:sen-ana} supplements the sensitivity analysis for Koopman operator $m$ and DSD threshold $\xi$. 
ReLaX is generally robust to $m$, except at $m=5$, where \textit{too few spectral modes fail to capture intrinsic latent dynamics accurately}. This leads to unstable DSD calculation, rendering the regularization ineffective and degenerating the method into vanilla GRPO with premature convergence.
As for $\xi$, removing this constraint ($\xi=\infty$) allows DSD to grow unconstrained, leading to numerical instability and training collapse after $\sim 50$ steps. In contrast, setting $\xi=10,25,50$ effectively bounds DSD, with comparable performance across values, indicating robustness to the choice of $\xi$.

\subsection{Computational Time Consumption}
Compared with vanilla GRPO, ReLaX introduces additional computation for the model's latent representation. To assess its time efficiency, we provide a detailed runtime breakdown of the extra components, including Koopman dictionary learning and the actor update, and compare them against the vanilla GRPO. Following our ablation setup, this analysis is conducted on Qwen2.5-3B-VL and Qwen2.5-7B-Math. 

As shown in Table~\ref{tab:runtime}, fitting the Koopman dictionary incurs only a small one-time cost (109 s for the 3B model and 132 s for the 7B model), which is negligible compared to the whole training process, as it is performed only once at initialization. During training, the primary additional overhead of ReLaX arises from computing the DSD score for each hidden state at every step, which increases the actor-update time by roughly 50\% compared with vanilla GRPO. However, this component accounts for only about 10\% of the total per-step runtime, making its impact on overall training time relatively minor. These observations indicate that the time overhead introduced by ReLaX is acceptable in practice, and we leave further optimization of its computational efficiency to future work.

\begin{table}[t]
\centering
\small
\begin{tabular}{lcc}
\toprule
\textbf{Component} & \textbf{GRPO (s)} & \textbf{ReLaX (s)} \\
\midrule
\multicolumn{3}{l}{\textbf{Qwen2.5-3B-VL}} \\
\midrule
Fit Koopman dictionary   & -   & 109 \\
Update actor   & 490   & 578 \\
\textbf{Total / step}  & 1052  & 1161 \textcolor{BrickRed}{(+10.4\%)} \\
\midrule
\multicolumn{3}{l}{\textbf{Qwen2.5-7B-Math}} \\
\midrule
Fit Koopman dictionary   & -    & 132 \\
Update actor    &   449       & 653 \\
\textbf{Total / step}  & 2030   & 2273 \textcolor{BrickRed}{(+12.0\%)} \\
\bottomrule
\end{tabular}
\caption{Comparison of the runtime breakdown per training step between vanilla GRPO and the proposed ReLaX on Qwen2.5-3B-VL and Qwen2.5-7B-Math.}
\label{tab:runtime}
\end{table}

\subsection{More Details on Comparisons with Token-level Methods}

To support the claims and conclusions presented in Sec.~\ref{sec:exp_compare}, we provide additional results from analytical experiments comparing ReLaX with token-level methods. The following section presents evaluation results across all multimodal reasoning benchmarks, as well as a detailed case study of the obtained LRMs.

\subsubsection{Extra Evaluation Results of VLMs}
Firstly, we provide additional results of token-level methods on 3B-VL models evaluated across all multimodal benchmarks. In addition to the results on MathVista, DynaMath, MMStar, and the two multidisciplinary subsets of EMMA reported in Fig.~\ref{fig:mllm-compare-a}, we further include comparisons on MathVerse, MathVision, MMMU, and the full EMMA benchmark in Fig.~\ref{fig:mllm-compare-benchmark-supp}. These results reinforce that while ReLaX consistently outperforms the vanilla GRPO baseline, its main advantage over the token-level method, KL-Cov, appears in the multidisciplinary multimodal benchmarks, which rely more on visual information, rather than in the mathematics-dominant benchmarks where text plays a central role. Additional training dynamics of the reward and gradient norm are presented in Fig.~\ref{fig:mllm-compare-dynamics-supp}. We observe that ReLaX exhibits a notably stable gradient norm, which is beneficial for optimization.

\subsubsection{Case Study for Multimodal Reasoning}
To investigate the behavior of models trained with the proposed ReLaX in multimodal reasoning, we present a detailed case study comparing ReLaX and KL-Cov~\cite{cui2025entropy} on 3B-VL models. We select a query from DynaMath~\cite{zou2025dynamath}, a recent benchmark designed to evaluate reasoning robustness by testing how well models perform under different variants of the same question, such as changes in visual numerical values. As shown in Table~\ref{tab:case-study-dynamath}, the question asks for the perimeter of a rectangular prism, and the three variants differ only in the numerical values of length, width, and height. ReLaX-3B-VL consistently follows the correct formula, $P = 4(L + W + H)$, and produces accurate answers across all variants. In contrast, the model trained with KL-Cov shows clear sensitivity to these variations. Specifically, in the second variant, it fails to extract the height from the image and incorrectly treats the rectangular prism as a flat rectangle. In the third variant, it applies an incorrect computation procedure for the perimeter. 
Additional support for ReLaX’s accurate visual understanding in multidisciplinary reasoning tasks is presented in Table~\ref{tab:case-study-MMMU-Physics}.

\subsubsection{Case Study for Text-only Reasoning}
We further analyze the qualitative results on AMC23 (Tab.~\ref{tab:case-study-amc}), comparing two 7B models fine-tuned from Qwen2.5-Math: one trained with ReLaX and the other with R1-zero-Div~\cite{yao2025diversity}. We examine two representative cases in which both models produce the correct final answer. Notably, both models engage in self-examination after generating the initial solution; however, this behavior reveals a fundamental difference in the quality of their reasoning.
The model trained by the proposed ReLaX re-examines its results by applying the law of cosines, a mathematically grounded verification step directly relevant to the problem context. In contrast, R1-zero-Div attempts to validate its answer by generating Python code—despite lacking any execution environment or interpreter. This constitutes a form of hallucination induced by coarse-grained entropy encouragement: increasing token-level entropy may elevate surface-level diversity but offers no guarantee that the generated content remains meaningful.

\clearpage
\begin{table*}[t]
\centering
\begin{tabularx}{\textwidth} { 
 >{\raggedright\arraybackslash}X  }
\textbf{Question:} In a table tennis tournament every participant played every other participant exactly once. Although there were twice as many right-handed players as left-handed players, the number of games won by left-handed players was $40\%$ more than the number of games won by right-handed players. (There were no ties and no ambidextrous players.) What is the total number of games played? \\
\hline
\textbf{ReLaX:}
To solve the problem, we start by letting the side lengths of triangle $ABC$ be $a$, $b$, and $c$ with $a$ being the smallest side and the sides in arithmetic progression. Since $a$ is the smallest side and has a length of $6$, we can let the side lengths be $6$, $6+d$, and $6+2d$. Without loss of generality, we assume that the angle opposite the largest side is $120°$. Thus, the largest side is $6+2d$.\\
\textbf{...} \\
The area $A$ of the triangle is given by: \\
$A = \sqrt{s(s-a)(s-b)(s-c)} = \sqrt{15(15-6)(15-10)(15-14)} = \sqrt{15 \cdot 9 \cdot 5 \cdot 1} = \sqrt{675} = 15\sqrt{3}$\\
\textcolor{OliveGreen}{However, we should verify our result using \textbf{the Law of Cosines} as well to ensure the correct angle is $120°$: } \\
$c^2 = a^2 + b^2 - 2ab \cos(C)$ \\
\textbf{...} \\
\textcolor{OliveGreen}{This confirms that our side lengths and angle are correct. Thus, the area of the triangle is: $15\sqrt{3}$} \\

The problem specifies that the final answer should be in the form $m\sqrt{n}$ where $m$ and $n$ are positive integers and $n$ has no square factors. Here, $m = 15$ and $n = 3$. Therefore, $m + n = 15 + 3 = 18$.\\
Thus, the final answer is: $\boxed{18}$\\
\textbf{[CORRECT]}\\
\hline
\textbf{R1-zero-Div:}
Given that the side lengths of triangle $ABC$ are in arithmetic progression and the smallest side has length $6$, let's denote the side lengths as $6$, $6+d$, and $6+2d$ for some positive number $d$. Without loss of generality, we can assume that the angle of $120°$ is opposite the largest side, which is $6+2d$. \\
By the Law of Cosines, we have: \\
$\left(6+2d\right)^2 = 6^2 + \left(6+d\right)^2 - 2 \cdot 6 \cdot \left(6+d\right) \cdot \cos\left(120°\right)$\\
\textbf{...} \\
Thus, $m = 15$ and $n = 3$, and the final answer is $m + n = 15 + 3 = 18$.
\textcolor{red}{Let's confirm this using \textbf{Python and SymPy}.} \\
\parbox[c]{1cm}{\centering \colorbox{Salmon}{ ```python}} \\
\parbox[c]{1cm}{\centering \colorbox{Salmon}{import sympy as sp}} \\
\parbox[c]{1cm}{\centering \colorbox{Salmon}{\textbf{...}}} \\
\parbox[c]{1cm}{\centering \colorbox{Salmon}{print(result)}} \\
\parbox[c]{1cm}{\centering \colorbox{Salmon}{```}}\\
\parbox[c]{1cm}{\centering \colorbox{Salmon}{Output: `18'}} \\
The area of triangle $ABC$ is $15\sqrt{3}$. Therefore, $m = 15$ and $n = 3$, and the final answer is $m + n = 18$. \\
Thus, the final answer is $\boxed{18}$. \\
\textbf{[CORRECT]}\\
\hline
\end{tabularx}
\caption{Comparison of model responses on AMC23 between ReLaX and R1-zero-Div. The self-verification behaviors exhibited during reasoning are highlighted in \textcolor{OliveGreen}{green} and \textcolor{red}{red}, respectively.}
\label{tab:case-study-amc}
\end{table*}

\clearpage
\begin{table*}[t]
    \centering
    \begin{tabularx}{.95\textwidth}{X|X|X}
    \multicolumn{3}{c}{
    \begin{minipage}{0.9\textwidth}
    \textbf{Question:} Given a rectangular prism with length $l$, width $w$, and height $h$, calculate its perimeter.
    \end{minipage}
    } \\
    \hline
    \small
    {
    \begin{minipage}[t]{.5\columnwidth}
    \centering
            {\includegraphics[width=\linewidth]{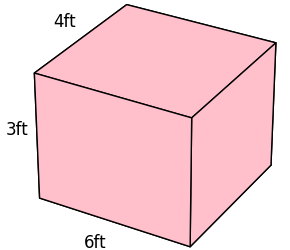}}\\
            Question Variant 1
    \end{minipage}} & {
    \begin{minipage}[t]{.5\columnwidth}
    \centering
            {\includegraphics[width=\linewidth]{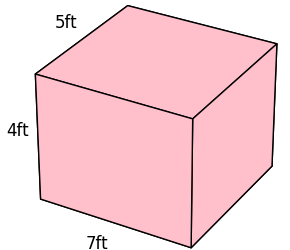}}\\
            Question Variant 2
    \end{minipage}} & {
    \begin{minipage}[t]{.5\columnwidth}
    \centering
            {\includegraphics[width=\linewidth]{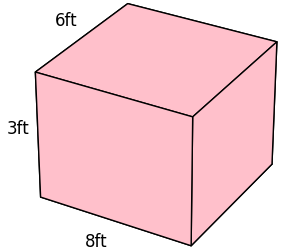}}\\
            Question Variant 3
    \end{minipage}}\\
    \hline
    {\begin{minipage}[t]{.6\columnwidth}
    \textbf{ReLaX:} To find the perimeter of a rectangular prism, ...
    
    Given:
    - Length (L) = 6 ft
    - Width (W) = 3 ft
    - Height (H) = 4 ft
    
    The formula for the perimeter \( P \) of a rectangular prism is:
    \parbox[c]{5cm}{\centering \colorbox{lime}{$ P = 4(L + W + H) $}}
    
    Substituting the given values into the formula:

    \parbox[c]{5cm}{\centering $ P = 4(6 + 3 + 4) $}
    \parbox[c]{5cm}{\centering$ P = 4(13) $}
    \parbox[c]{5cm}{\centering$ P = 52 $}
    
    So, the perimeter of the rectangular prism is \boxed{52} feet.
    \textbf{[CORRECT]}
    \end{minipage}}
    &
    {\begin{minipage}[t]{.6\columnwidth}
    \textbf{ReLaX:} To find the perimeter of a rectangular prism, ...
    
    Given:
    - Length (L) = 7 ft
    - Width (W) = 4 ft
    - Height (H) = 5 ft
    
    The formula for the perimeter \( P \) of a rectangular prism is:
    \parbox[c]{5cm}{\centering \colorbox{lime}{$ P = 4(L + W + H) $}}
    
    Substituting the given values into the formula:

    \parbox[c]{5cm}{\centering $ P = 4(7 + 4 + 5) $}
    \parbox[c]{5cm}{\centering$ P = 4(16) $}
    \parbox[c]{5cm}{\centering$ P = 64 $}
    
    So, the perimeter of the rectangular prism is \boxed{64} feet.
    \textbf{[CORRECT]}
    \end{minipage}}
    &
    {\begin{minipage}[t]{.6\columnwidth}
    \textbf{ReLaX:} To find the perimeter of a rectangular prism, ...
    
    Given:
    - Length (L) = 8 ft
    - Width (W) = 3 ft
    - Height (H) = 6 ft
    
    The formula for the perimeter \( P \) of a rectangular prism is:
    \parbox[c]{5cm}{\centering \colorbox{lime}{$ P = 4(L + W + H) $}}
    
    Substituting the given values into the formula:
    
    \parbox[c]{5cm}{\centering $ P = 4(8 + 3 + 6) $}
    \parbox[c]{5cm}{\centering$ P = 4(17) $}
    \parbox[c]{5cm}{\centering$ P = 68 $}
    
    So, the perimeter of the rectangular prism is \boxed{68} feet. \textbf{[CORRECT]}
    \end{minipage}}
    \\
    \hline
    {\begin{minipage}[t]{.6\columnwidth}
    \textbf{KL-Cov:}  To find the perimeter of the rectangular prism, ...
    
    The formula for the perimeter of the edges of a rectangular prism is:
    \parbox[c]{5cm}{\centering \colorbox{lime}{$ P = 4(l + w + h) $}}
    where \( l \), \( w \), and \( h \) are the length, width, and height of the prism, respectively. Substituting the given dimensions (6 feet, 3 feet, 4 feet):
    \[ P = 4(6 + 3 + 4) = 4 \times 13 = 52 \text{ feet} \]
    
    However, the problem seems to be asking for the perimeter of the faces, which is not a standard geometric term either. Given the dimensions provided, if we assume the problem is asking for the perimeter of the faces that would form a net of the prism, we need to consider the perimeter of the faces, ...
    
    Therefore, the correct answer is:
    \( \boxed{52} \) 
    \textbf{[CORRECT]}
    \end{minipage}}
    &
    {\begin{minipage}[t]{.6\columnwidth}
    \textbf{KL-Cov:}  To find the perimeter of the rectangular prism, ...
    
    \textbf{\textcolor{red}{The dimensions given for the base of the rectangular prism are 7 feet and 4 feet.}} The perimeter \(P\) of a rectangle is calculated using the formula:
    \parbox[c]{5cm}{\centering \colorbox{Salmon}{$ P = 2 \times (length + width) $}}
    Here, the length is 7 feet and the width is 4 feet. Substituting these values into the formula, we get:
    \[ P = 2 \times (7 + 4) = 2 \times 11 = 22 \text{ feet} \]
    
    So, the perimeter of the rectangular prism (interpreted as the perimeter of the base) is:
    \( \boxed{22} \)
    \textbf{[WRONG]}
    \end{minipage}}
    &
    {\begin{minipage}[t]{.6\columnwidth}
    \textbf{KL-Cov:}  To find the perimeter of the rectangular prism, \textbf{\textcolor{violet}{we need to calculate the perimeter of each pair of opposite faces and then sum them up}}.
    
    The perimeter of each pair of opposite faces is calculated as follows:
    1. For the faces with dimensions 8 feet by 3 feet, the perimeter of one face is \(2 \times (8 + 3) = 2 \times 11 = 22\) feet. Since there are two such faces, the total contribution to the perimeter from these faces is \(2 \times 22 = 44\) feet.
    2. ..., the total contribution to the perimeter from these faces is \(2 \times 28 = 56\) feet.
    3. ..., the total contribution to the perimeter from these faces is \(2 \times 18 = 36\) feet.
    
    Adding these contributions together gives us the total perimeter of the rectangular prism:
    \colorbox{Thistle}{$44 + 56 + 36 = 136$}
    
    Therefore, the perimeter of the rectangular prism is \(\boxed{136}\). 
    \textbf{[WRONG]}
    \end{minipage}}
    \\
    \hline
    \end{tabularx}
    \caption{Comparison of model responses on DynaMath between ReLaX and KL-Cov using three query variants with modified visual content, all paired with the same text prompt: “\textbf{Find the perimeter of the rectangular prism.}” Two typical reasoning failures caused by KL-Cov are highlighted in \textcolor{red}{red} and \textcolor{violet}{purple}, respectively, while the correct counterparts by ReLaX are highlighted in \colorbox{lime}{green}.}
    \label{tab:case-study-dynamath}
\end{table*}

\clearpage
\begin{table*}[t]
    \centering
    \begin{tabularx}{.95\textwidth}{X|X}
    \multicolumn{2}{c}{
    \begin{minipage}{0.9\textwidth}
    \textbf{Question:} A light ray enters a block of plastic and travels along the path shown. By considering the behavior of the ray at point $P$, determine the speed of light in the plastic. Options:  A. 0.44 ;B. 0.88; C. 1.13; D. 2.26.\\
    \end{minipage}
    } \\
    \multicolumn{2}{c}{
    \begin{minipage}[t]{.5\columnwidth}
    \centering
            {\includegraphics[width=\linewidth]{figures/supp-MMMU-phys.jpg}}
    \end{minipage}} \\
    \hline
    {\begin{minipage}[t]{.95\columnwidth}
    \textbf{ReLaX:} To determine the speed of light in the plastic, we need to use Snell's Law,...

\[ n_1 \sin(\theta_1) = n_2 \sin(\theta_2) \]

Given: \\
- \colorbox{lime}{The angle of incidence \( \theta_1 = 53^\circ \).} \\
- \colorbox{lime}{The angle of refraction \( \theta_2 = 37^\circ \).}

We know that the index of refraction for air is approximately \( n_1 = 1 \). We need to find the index of refraction for the plastic, \( n_2 \).

Using Snell's Law:

\[ 1 \cdot \sin(53^\circ) = n_2 \cdot \sin(37^\circ) \]
...

The speed of light in a medium is given by:

\[ v = \frac{c}{n} \]

...

Therefore, the speed of light in the plastic is:

\[ v = \frac{3 \times 10^8 \text{ m/s}}{1.33} \approx 2.26 \times 10^8 \text{ m/s} \]

Hence, the correct answer is:

\[ \boxed{D} \]
    \textbf{[CORRECT]}
    \end{minipage}}
    &
    {\begin{minipage}[t]{.95\columnwidth}
    \textbf{KL-Cov:} To solve this problem, we need to use the principles of refraction and Snell's Law. \colorbox{Salmon}{We are given the angle of incidence at point \(P\) as \(37^\circ\) }\\
    \colorbox{Salmon}{and the angle of refraction inside the plastic as \(53^\circ\)}.

First, let's recall Snell's Law, which states:
\[ n_1 \sin \theta_1 = n_2 \sin \theta_2 \]
...
The refractive index of the plastic is \(n_2 = 0.75\). The speed of light in a medium is given by:
\[ v = \frac{c}{n} \]

However, we need to check the options provided. The options are in terms of a fraction of \(10^8 \, \text{m/s}\). The correct value for the refractive index of the plastic is actually \(n_2 = \frac{3}{4} = 0.75\), but we need to re-evaluate the options given. The correct option based on the refractive index calculation is:

\[ v = \frac{3 \times 10^8 \, \text{m/s}}{0.75} = 4 \times 10^8 \, \text{m/s} \approx 0.88 \times 10^8 \, \text{m/s} \]

So the correct answer is:
\[ \boxed{B} \]
    \textbf{[WRONG]}
    \end{minipage}}    \\
    \hline
    \end{tabularx}
    \caption{Comparison of model responses on MMMU between ReLaX and KL-Cov. Failures of KL-Cov in correctly interpreting the visual input are highlighted in \colorbox{Salmon}{red}, whereas the correct responses produced by ReLaX are highlighted in \colorbox{lime}{green}.}
    \label{tab:case-study-MMMU-Physics}
\end{table*}